\documentclass{naturep}

\usepackage{scicite}
\usepackage{times}

\usepackage{amssymb}
\usepackage{amsmath}
\usepackage{graphicx}
\usepackage{algorithmicx}
\usepackage{algorithm}

\usepackage[noend]{algpseudocode}
\usepackage{bbm}
\usepackage[margin=0.6in]{geometry}
\usepackage{ragged2e}
\usepackage{setspace}
\usepackage{longtable}
\usepackage{anyfontsize}
\usepackage{setspace}
\usepackage{float}
\usepackage{booktabs}
\usepackage{multirow}
\usepackage{xcolor}
\usepackage[colorlinks = true,
            linkcolor = blue,
            urlcolor  = blue,
            citecolor = blue,
            anchorcolor = blue]{hyperref}

\usepackage{caption}
\usepackage{subcaption}

\makeatletter
\let\saved@includegraphics\includegraphics
\AtBeginDocument{\let\includegraphics\saved@includegraphics}

\makeatother

\newcommand{\rjc}[1]{{\textcolor[rgb]{0,0,0}{#1}}}

\title{\begin{flushleft}{\begin{spacing}{1}Pan-Cancer Integrative Histology-Genomic Analysis via Interpretable Multimodal Deep Learning\end{spacing}}\end{flushleft}}

\begin{document}

\maketitle
\begin{spacing}{1.5}
\vspace{-15mm}
\noindent Richard J. Chen$^{1,2,3,4}$, Ming Y. Lu$^{1,3,4,\ddag}$, Drew F. K. Williamson$^{1,3,\ddag}$, Tiffany Y. Chen$^{1,3,\ddag}$, Jana Lipkova$^{1,3,4}$, Muhammad Shaban$^{1,3,4}$, Maha Shady$^{1,2,3}$,  Mane Williams$^{1,2,3}$, Bumjin Joo$^{1}$, Zahra Noor$^{5}$, and Faisal Mahmood$^{*1,3,4}$
\begin{affiliations}
 \item Department of Pathology, Brigham and Women's Hospital, Harvard Medical School, Boston, MA
 \item Department of Biomedical Informatics, Harvard Medical School, Boston, MA
 \item Cancer Program, Broad Institute of Harvard and MIT, Cambridge, MA 
 \item Cancer Data Science Program, Dana-Farber Cancer Institute, Boston, MA
 \item Department of Computer Science, Harvard University, Cambridge, MA
 \\$\ddag$ Contributed Equally
\end{affiliations}
 

\noindent\textbf{Interactive Web Database:} \href{http://pancancer.mahmoodlab.org}{http://pancancer.mahmoodlab.org}\\
\noindent\textbf{Open-source Code:} \href{https://github.com/mahmoodlab/PORPOISE}{https://github.com/mahmoodlab/PORPOISE}\\

\end{spacing}
\begin{spacing}{1.3}
\noindent\textbf{*Correspondence:}\\ 
Faisal Mahmood \\
60 Fenwood Road, Hale Building for Transformative Medicine\\
Brigham and Women's Hospital, Harvard Medical School\\
Boston, MA 02445\\
faisalmahmood@bwh.harvard.edu
\end{spacing}

\newpage
\noindent\textbf{\large{Abstract}}
\vspace{-10mm}
\begin{spacing}{1}
\noindent 




\noindent The rapidly emerging field of deep learning-based computational pathology has demonstrated promise in developing objective prognostic models from histology whole slide images. However, most prognostic models are either based on histology or genomics alone and do not address how histology and genomics can be integrated to develop joint image-omic prognostic models. Additionally identifying explainable morphological and molecular descriptors from these models that govern such prognosis is of interest. We used multimodal deep learning to integrate gigapixel whole slide pathology images, RNA-seq abundance, copy number variation, and mutation data from 5,720 patients across 14 major cancer types. Our interpretable, weakly-supervised, multimodal deep learning algorithm is able to fuse these heterogeneous modalities for predicting outcomes and discover prognostic features from these modalities that corroborate with poor and favorable outcomes via multimodal interpretability. We compared our model with unimodal deep learning models trained on histology slides and molecular profiles alone, and demonstrate performance increase in risk stratification on 9 out of 14 cancers. In addition, we analyze morphologic and molecular markers responsible for prognostic predictions across all cancer types. All analyzed data, including morphological and molecular correlates of patient prognosis across the 14 cancer types at a disease and patient level are presented in an interactive open-access database (\href{http://pancancer.mahmoodlab.org}{http://pancancer.mahmoodlab.org}) to allow for further exploration and prognostic biomarker discovery. To validate that these model explanations are prognostic, we further analyzed high attention morphological regions in WSIs, which indicates that tumor-infiltrating lymphocyte presence corroborates with favorable cancer prognosis on 9 out of 14 cancer types studied.


\end{spacing}

\newpage
\begin{spacing}{1}

\vspace{-4mm}

The recent advancements made in deep learning for computational pathology have enabled the use of whole slide images (WSIs) for automated cancer diagnosis and objective characterization of morphological phenotypes in the tumor microenvironment\cite{lecun2015deep}. Using weakly-supervised deep learning, current state-of-the-art methods are able to harness large-scale WSI repositories and train on thousands of gigapixel-sized resection slides without requiring detailed pixel-level annotations from pathologists\cite{campanella2019clinical}. Recent studies have shown that using only image-level labels such as cancer grade and tissue type, deep learning is able to achieve remarkable performance on tasks such as lymph node metastasis detection and cancer origin prediction in tumors of unknown primary\cite{lu2019semi, lu2021data, lu2021ai}. After training and inference, interpretability techniques have emerged that can visualize morphological regions of high diagnostic relevance (high attention) made by the model\cite{ilse2018attention}. In addition to cancer diagnosis, deep learning models trained with patient outcome information have also been shown to stratify patients into relevant risks groups in cancer prognosis\cite{chen2020pathomic}. By characterizing high attention morphological regions from models supervised with survival outcome labels, there is enormous potential in using deep learning for automated biomarker discovery of novel and prognostic morphological determinants\cite{beck2011systematic, echle2020deep, diao2021human}.

\vspace{-3mm}
Despite the progress made in computational pathology, cancer prognostication is a difficult task that is driven by markers in both histology and genomics. With the advent of modern sequencing technologies, the classification and prognostication of many tumors has evolved to incorporate genomics and transcriptomics information in addition to histology-based analyses of the tumor microenvironment\cite{louis2016who,galateau2016who}. However, current deep learning approaches for cancer prognosis suffer from limitations in that they: 1) are not multimodal in integrating histology and genomics data in an intuitive manner, 2) require pathologist guidance and/or selective sampling of representative ROIs from tumor regions, or 3) are not interpretable in identifying and characterizing prognostic morphological and molecular features\cite{yu2016predicting,yousefi2017predicting, mobadersany2018predicting, courtiol2019deep, wulczyn2020deep, lu2020federated, wang2019weakly, chen2021whole}. Recent pan-cancer studies on the TCGA have mainly focused on learning genotype-phenotype associations via predicting molecular features from histopathology region-of-interests\cite{fu2020pan, kather2020pan}. However, these approaches capture only existing shared information between histology and genomics, and do not address how histology and genomics can be combined for discovering novel prognostic biomarkers.

\vspace{-5mm}

\section*{\large{Results}}
\vspace{-4mm}

\section*{Interpretable, Multimodal Deep Learning for Knowledge Discovery}
\vspace{-4mm}

\begin{figure*}[h]
\vspace{-12mm}
\includegraphics[width=\textwidth]{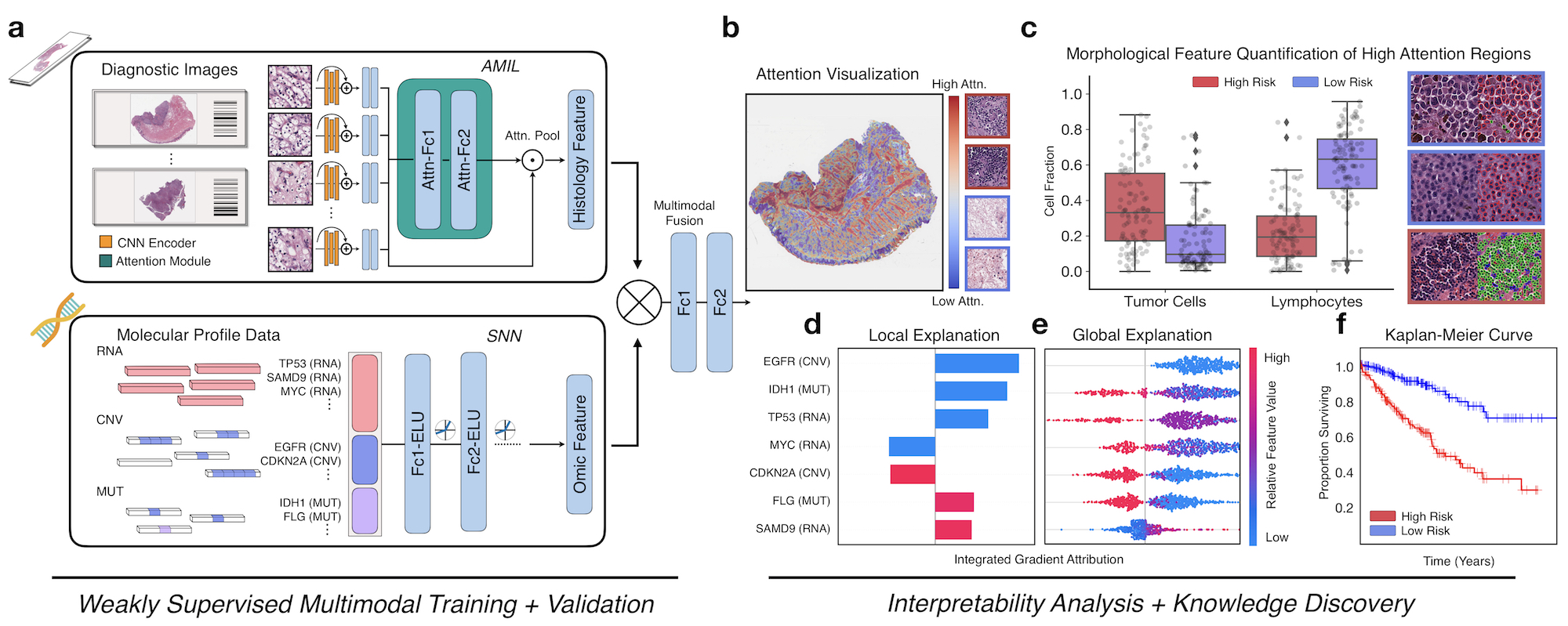}
\caption*{\textbf{Fig. 1: Pathology-Omic Research Platform for Integrative Survival Estimation (PORPOISE) Workflow.} \textbf{a.} Patient data in the form of digitized high-resolution FFPE H\&E histology glass slides (known as WSIs) with corresponding molecular data are used as input in our algorithm. Our multimodal algorithm consists of three neural network modules trained end-to-end: 1) an attention-based multiple instance learning (AMIL) network for processing WSIs, 2) a self-normalizing network (SNN) for processing molecular data features, and 3) a multimodal fusion layer that computes the Kronecker Product to model pairwise feature interactions between histology and molecular features. Following inference and training, local- and global-level interpretability is performed to understand how WSI and molecular features impact cancer prognosis. \textbf{b.} For WSIs, per-patient local explanations are visualized as high-resolution attention heatmaps using attention-based interpretability, in which high attention regions (red) in the heatmap correspond to morphological features that contribute to the model's predicted risk score. \textbf{c.} Global morphological patterns are extracted via cell quantification of high attention regions in low and high risk patient cohorts. \textbf{d.} For molecular features, per-patient local explanations are visualized using attribution-based interpretability, in which attribution color corresponds to low (blue) vs. high (red) gene feature value, attribution direction to how the gene feature impacts low risk (left) vs. high (risk), and attribution magnitude refers to gene feature importance for that individual patient. \textbf{e.} Global interpretability for molecular features is performed via analyzing the directionality, feature value and magnitude of gene attributions across all patients. \textbf{f.} Kaplan-Meier analysis is performed to visualize patient stratifcation of low and high risk patients for individual cancer types.
}
\end{figure*}

In order to address the difficulties in developing joint image-omic biomarkers that can be used for cancer prognosis, we propose a multimodal fusion deep learning-based algorithm that uses both scanned H\&E whole slide images and molecular profile features (mutation status, copy number variation, RNA-Seq expression) to measure and explain relative risk of cancer death (\textbf{Fig. 1a, Fig. 1f}). Our multimodal network is capable of not only integrating these two modalities in weakly-supervised learning tasks such as survival outcome prediction, but also explaining how histopathology features, molecular features, and their interactions contribute locally towards low- and high-risk outcomes in a specific patient. After risk assessment within a patient cohort, our network uses both attention- and attribution-based interpretability as an untargeted approach for estimating prognostic markers across all patients (\textbf{Fig. 1a-e}). Our study uses 6,592 WSIs from 5,720 patient samples across 14 cancer types from the TCGA, where multiple diagnostic slides for a single patient are treated as a single data point. For each cancer type, we trained our multimodal model in a 5-fold cross-validation using our weakly-supervised paradigm, and conducted ablation analyses comparing with the performance of unimodal models. Following training and model evaluation, we conducted extensive analyses on the interpretability of our networks, investigating not only local- and global-level image-omic explanations for each model within each cancer type, but also shifts in feature importance when comparing unimodal interpretability versus multimodal interpretability. 

Our ultimate contribution in this work is the development of a research tool that uses model explanations of both whole slide image and molecular profile data to drive the discovery of new prognostic biomarkers. Using our multimodal network, we developed the Pathology-Omics Research Platform for Integrative Survival Estimation (PORPOISE), an interactive, freely available platform that directly yields prognostic markers learned by our model for thousands of patients across our 14 cancer types, available at \href{http://pancancer.mahmoodlab.org}{http://pancancer.mahmoodlab.org} (\textbf{Interactive Demo}). Specifically, PORPOISE allows the user to visualize: 1) raw H\&E image pyramidal TIFFs overlayed with attention-based interpretability from both unimodal and multimodal training, 2) local explanations of molecular features using attribution-based interpretability for each patient, and 3) global patterns of morphological and molecular feature importance for each study. To validate that PORPOISE can be used to drive the discovery of joint image-omic biomarkers, we analyzed high attention morphological regions in WSIs to test the hypothesis that tumor-infiltrating lymphocyte presence corroborates with favorable cancer prognosis. \rjc{Unlike previous studies that have corroborated TIL with outcomes, our model identifies these morphological features on its own as being prognostic.}



To build PORPOISE, we adapted two recently described algorithms for rapid WSI processing and histology-omic fusion respectively\cite{lu2020data, chen2020pathomic}. The multimodal network in PORPOISE is composed of three subnetwork components: 1) an attention-based multiple instance learning network (AMIL) used for WSI inputs, 2) a self-normalizing network (SNN) for molecular feature components, and 3) and a multimodal fusion layer (MMF) for integrating feature representations from AMIL and SNN. For processing WSIs, in contrast with previous work that selectively sample patches from WSIs and matches the patch-level data with its patient-level label, our approach utilizes the entire tissue microenvironment across all diagnostic slides for both training and inference. Using only time-to-event patient death as the label, AMIL is able to localize (or attend to) morphological regions that have high prognostic relevance in the slides, and then aggregates these regions into a single low-dimensional feature representation. To process molecular features, we use SNN to learn a transformation of molecular data (containing vectorized gene mutation, copy number variation, and RNA-Seq features) into another low-dimensional feature representation. Following feature extraction using the AMIL and SNN subnetworks, our model constructs a joint multimodal feature representation that models pairwise interactions between histology and molecular features, which is then used to estimate the hazard function in survival analysis. In order to use entire WSIs as input into our model, we adapted the Cox proportional likelihood loss commonly used in training deep survival models to allow supervision with tiny batch sizes. Finally, to assess model explanations and interpretability, we applied both attention- and gradient-based interpretability to the WSI and molecular feature inputs respectively. For WSIs, attention scores are calculated for every $256 \times 256$ image patch within the tissue region, which are then used to visualize high-resolution attention heatmaps for human interpretability and validation\cite{ilse2018attention}. For molecular features, we used Integrated Gradients (IG), a model-agnostic gradient-based feature attribution method that attributes the risk prediction made by the model with respect to the molecular feature inputs\cite{sundararajan2017axiomatic}. Together, the combination of these interpretability techniques allows us to visualize and explain histology and molecular features used in our multimodal network, which we subsequently use to characterize prognostic morphological and molecular biomarkers and their impact in cancer prognosis. Further details of the model architecture, training, and dataset are described in the \textbf{Online Methods}.

\begin{figure*}[h]
\vspace{-10mm}
\includegraphics[width=\textwidth]{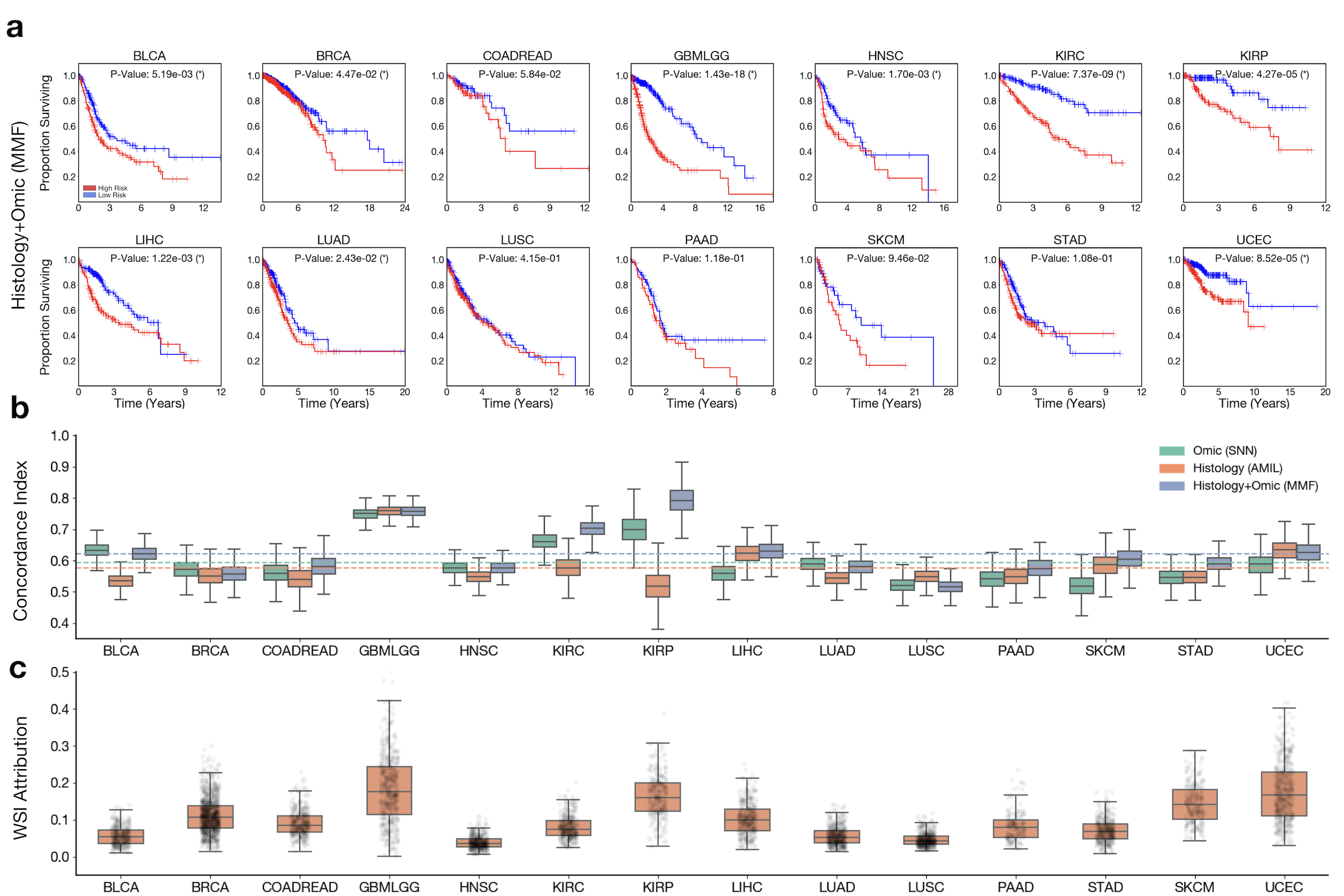}
\caption*{\textbf{Fig. 2: Model performances of PORPOISE and understanding impact of multimodal training.} \textbf{a.} Kaplan-Meier analysis of patient stratification of low and high risk patients via MMF across all 14 cancer types. Low and high risks are defined by the median 50\% percentile of hazard predictions via MMF. Logrank test was used to test for statistical significance in survival distributions between low and high risk patients (with * marked if P-Value $<$ 0.05). Kaplan-Meier analysis for AMIL and SNN are shown in \textbf{Fig. S11}. \textbf{b.} c-Index performance of SNN, AMIL and MMF in each cancer type in a 5-fold cross-validation (n=5,720). Horizontal line for each model shows average c-Index performance across all cancer types. Box plots correspond to c-indices of bootstrap replicates on the out-of-sample risk predictions in the validation folds. \textbf{c.} Distribution of WSI attribution across 14 cancer types. Each dot represents the proportion of feature attribution given to the WSI modality input compared to molecular feature input.
}
\vspace{-2mm}
\end{figure*}

\section*{Evaluation of Model Performance}
We first evaluated our proposed deep learning model for multimodal fusion (MMF) using 5-fold cross-validation on the paired WSI-molecular datasets of our 14 cancer type, in which each dataset was split into 5 80/20 partitions for training and validation. Using the same 5-fold cross-validation splits, we also compared our model with unimodal deep learning models trained with one modality - an AMIL model that uses only WSIs, and a SNN model that uses only molecular features. To compare the performances of these models, we used the cross-validated concordance index (c-Index) to measure the predictive performance of each model, Kaplan-Meier curves to visualize the quality of patient stratification between predicted high risk and low risk patient populations, and the logrank test to test statistical difference between high risk and low risk groups (\textbf{Fig. 2a}, \textbf{Fig. 2b}, \textbf{Table S3}). Across the 14 cancer types, MMF achieves an overall c-Index of 0.630, whereas AMIL and SNN have overall c-Indices of 0.585 and 0.597 respectively. In one-versus-all model performance comparisons on individual cancer types, MMF achieves the highest c-Index on 9 out of 14 (9/14) cancer types, whereas AMIL and SNN have the highest c-Index on 3/14 and 2/14 cancer types respectively. In addition to c-Index performance, MMF achieves statistical significance between predicted high and low risk patient populations on 9/14 cancer types, whereas AMIL and SNN are only statistically significant in 6/14 and 5/14 cancer types respectively. In head-to-head comparisons between the unimodal models, AMIL and SNN evenly outperforms each other on 7/14 cancer types, which suggests that for certain cancers such as LUSC, PAAD and UCEC, cancer prognosis can be better determined by histology rather than molecular features. Overall, though MMF does not always achieve the highest c-Index performance on every cancer type, the variability in c-Index performances and statistically significant patient stratification results from unimodal deep learning suggests that histology or molecular features alone are unable to generalize in explaining patient outcomes for all cancer types.

Amongst all cancer types included in our study, papillary renal cell carcinoma (KIRP) had the largest performance increase with multimodal training, reaching a c-Index performance of 0.811 (95\% CI: 0.697-0.874, P-Value = $4.27\times 10^{-5}$, logrank test), compared to 0.539 (95\% CI: 0.408-0.625, P-Value = $5.86\times 10^{-1}$, logrank test) using AMIL and 0.707 (95\% CI: 0.599-0.793, P-Value = $1.00\times 10^{-7}$, logrank test) using SNN (\textbf{Table S13}). Clear cell renal cell carcinoma (KIRC) showed similar improvement with multimodal training,  with a c-Index of 0.711 (95\% CI: 0.648-0.757, P-Value = $7.37\times 10^{-9}$, logrank test), compared to 0.567 (95\% CI: 0.508-0.650, P-Value = $5.98\times 10^{-2}$, logrank test) using AMIL and 0.665 (95\% CI: 0.607-0.721, P-Value = $2.34\times 10^{-5}$, logrank test) using SNN (\textbf{Table S13}). In visualizing the Kaplan-Meier survival curves of predicted high risk and low risk patient populations, MMF was observed to also show better discrimination of the two risk groups compared to SNN and AMIL (\textbf{Fig. 3c}, \textbf{Fig. 3i}). Furthermore, in quantifying the contribution of using WSIs in cancer prognosis, WSIs on average accounted for 8.9\% and 16.5\% of input attributions in MMF for KIRC and KIRP respectively, which suggests that molecular features drives most of the risk prediction in MMF (\textbf{Fig. 2c}). This corroborates with observations that molecular profiles are more prognostic for survival than WSIs in these two cancer types (in comparing the performances of SNN and AMIL), however, the inclusion of the WSIs in MMF still leads to significant improvement in the model's ability to stratify patients.  

In addition to conducting ablation studies in comparing unimodal and multimodal models, we also assessed Cox proportional hazard models using age, gender, and tumor grade covariates as baselines, which were still outperformed by MMF (\textbf{Table S4}). In head-to-head comparisons on cancer types with only available grade information, AMIL outperforms Cox models with an average c-Index of 0.601 compared to 0.592, which suggests that current subjective assessments for tumor grade in cancer prognosis can be refined using objective, deep learning-based phenotyping for evaluating prognosis.

\section*{Multimodal Interpretability for Joint Image-Omic Biomarker Discovery}

\begin{figure*}
\vspace{-12mm}
\includegraphics[width=\textwidth]{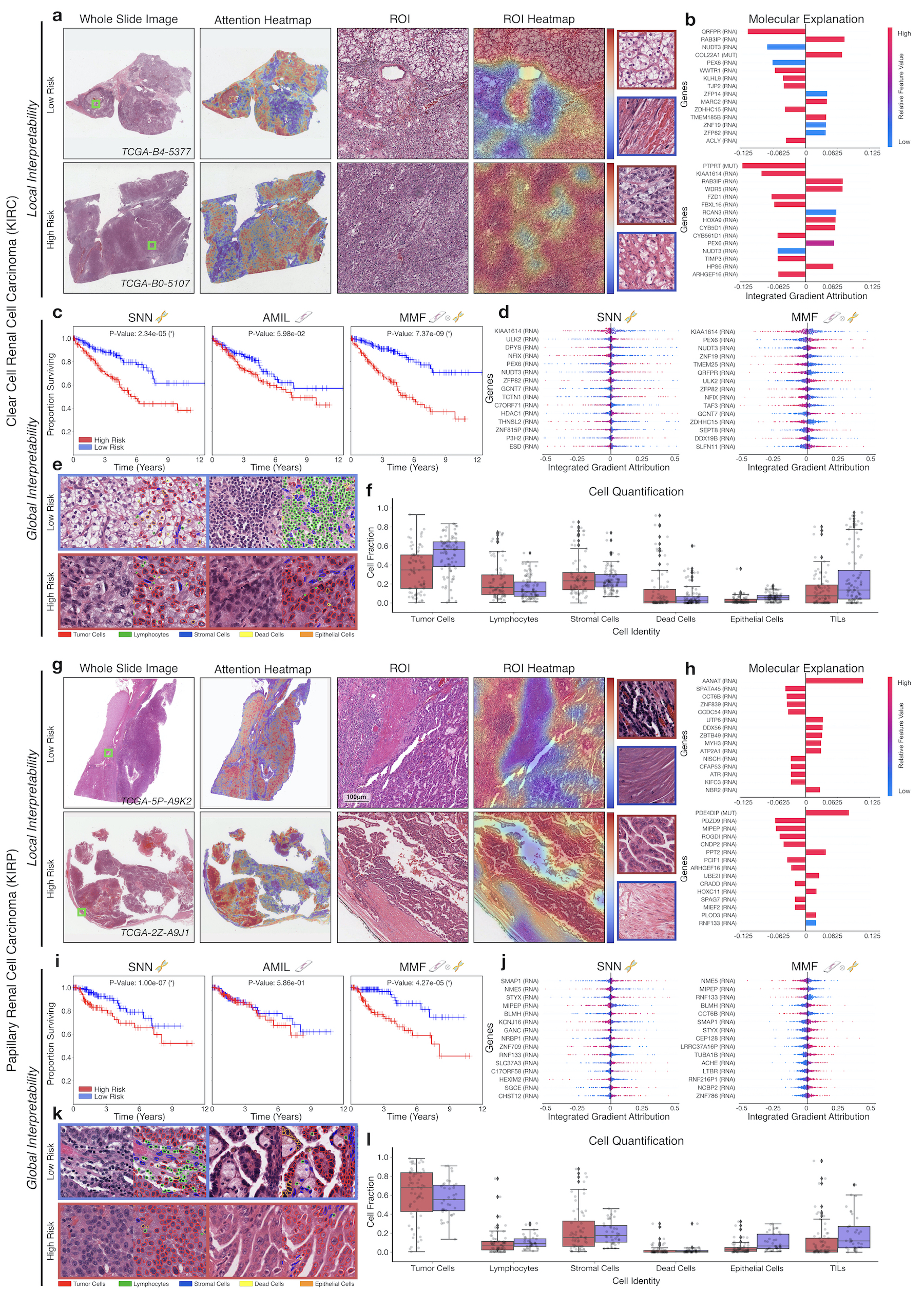}
\end{figure*}
\begin{figure*}
\caption*{\textbf{Fig. 3: Quantitative performance, local model explanation, and global interpretability analyses of PORPOISE on both clear cell and papillary renal cell carcinoma.} \textbf{a, g)} WSIs, associated attention heatmaps, ROIs, ROI heatmaps, and selected high-attention patches from example low-risk (top) and high-risk (bottom) cases, demonstrating areas to which PORPOISE assigns high attention for survival prediction. For KIRC (\textbf{a}), high attention for low-risk cases tends to focus on classic clear cell morphology while in high-risk cases, high attention often corresponds to areas with decreased cytoplasm or increased nuclear to cytoplasmic ratio. In both, high attention is paid to the renal capsule. For KIRP (\textbf{g}), low-risk cases often have high attention paid to complex and curving papillary architecture while for high-risk cases, high attention is paid to denser areas of tumor cells.  \textbf{b, h)} Local gene attributions for the corresponding low-risk (top) and high-risk (bottom) cases from a and g. \textbf{c, i)} Kaplan–Meier curves for omics-only (left, "SNN"), histology-only (center, "AMIL") and multimodal fusion (right, "MMF"), showing improved separation using MMF. \textbf{d, j)} \rjc{Global gene attributions across patient cohorts according to unimodal interpretability (left, "SNN"), and multimodal interpretability (right, "MMF"), with gene RNA-Seq abundances receiving the most attribution. SNN and MMF were both able to identify immune-related and prognostic markers such as NFIX and ULK in KIRC, and NME5 and MIPEP in KIRP. MMF additionally attributes to other immune-related / prognostic genes such as TMEM25, TAF3, TUBA1B, and ACHE.} \textbf{e, h)} Exemplar high attention patches from low-risk (top) and high-risk (bottom) cases with corresponding cell labels. \textbf{f, i)} Quantification of cell types in high-attention patches for each disease overall, showing increased TIL presence in both cancers.}
\end{figure*}

For interpretability and further validation of our models, we applied attention- and gradient-based interpretability to our trained SNN, AMIL, and MMF models in order to explain how WSI and molecular features are respectively used to predict risk of cancer death. For WSIs, we created a custom visualization tool that overlays attention weights computed from AMIL (and the AMIL subnetwork from MMF) onto the H\&E diagnostic slide, which is displayed as a high-resolution attention heatmap (\textbf{Interactive Demo}). For molecular features, we used SHAP-styled attribution decision plots to visualize the attribution weight and direction of each molecular feature calculated by Integrated Gradients in SNN (and the SNN subnetwork of MMF) (\textbf{Interactive Demo}). These interpretability and visualization techniques were then used to build our discovery platform, PORPOISE, which we then applied to each of our models and across all patients in the validation splits of our 5-fold cross-validation, yielding attention heatmaps and attribution decision plots for all 6,592 WSIs and 5,720 patients. \rjc{We did not include age and gender in our models, as we wanted our model predictions to be driven by only histology and molecular feature inputs for biomarker discovery.} Visualizations for analyses for individual patient model explanations in PORPOISE are termed local interpretability, with analyses performed on cancer-wide patient populations termed global interpretability. \textbf{Fig. 2} and \textbf{Fig. 3} show local and global interpretability for KIRC, KIRP, HNSC, and SKCM. Global interpretability results for the rest of the cancer types, as well as local interpretability results for all patients in the best validation splits are made available in PORPOISE.

\begin{figure*}
\vspace{-12mm}
\includegraphics[width=\textwidth]{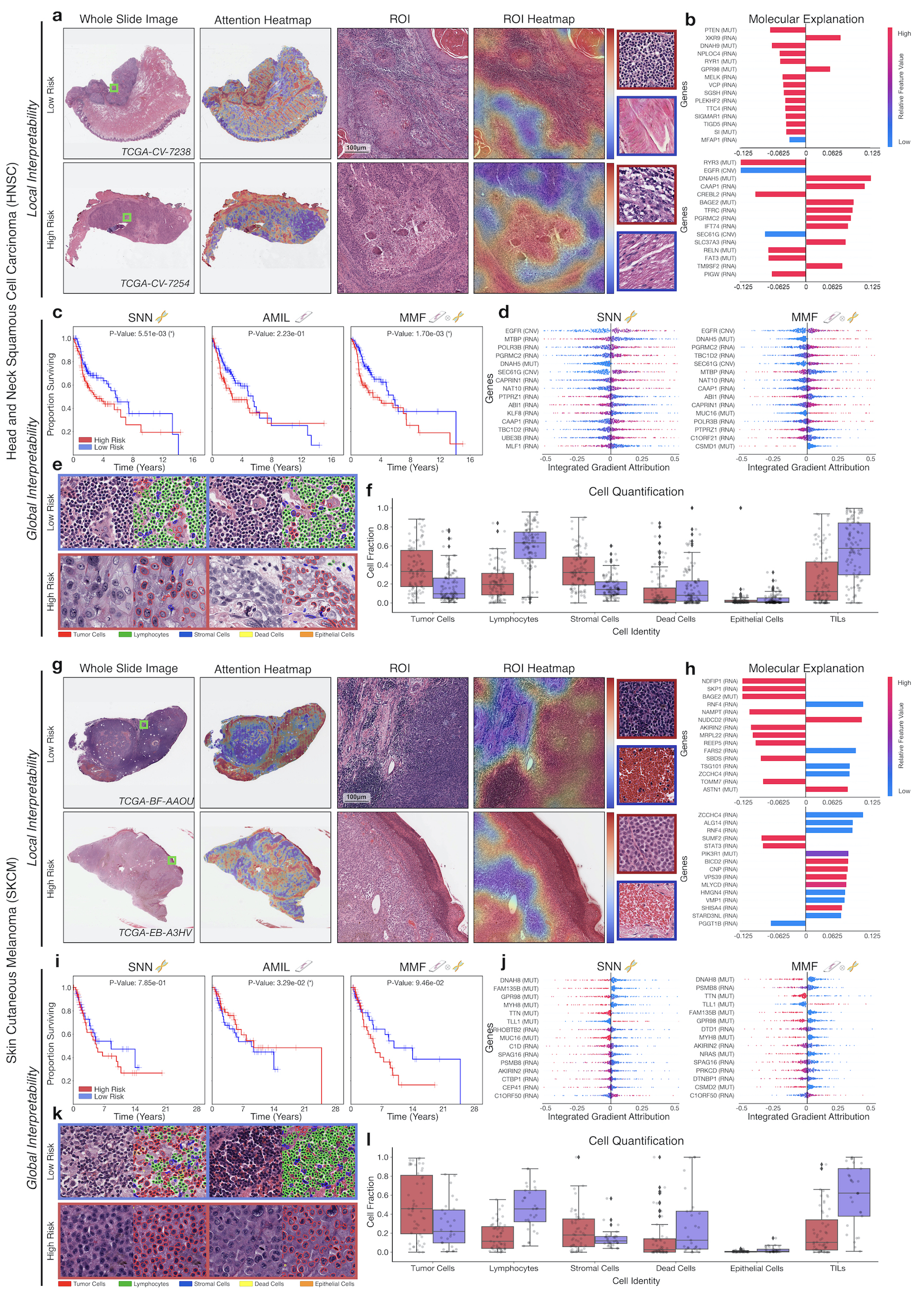}
\end{figure*}
\begin{figure*}
  \caption*{\textbf{Fig. 4: Quantitative performance, local model explanation, and global interpretability analyses of PORPOISE on both head and neck squamous cell carcinoma and skin cutaneous melanoma.} \textbf{a,g)} WSIs, associated heatmaps, ROIs, ROI heatmaps, and selected high-attention patches from example low-risk (top) and high-risk (bottom) cases, demonstrating areas to which PORPOISE assigns high attention for survival prediction. For head and neck squamous cell carcinoma (\textbf{a}), high attention for low-risk cases tends to focus on regions with increased tumor infiltrating lymphocytes, while in high-risk cases, high attention areas corresponded with regions with central necrosis. For both high and low-risk cases, the low attention regions focused on mainly background stroma. For melanoma (\textbf{g}), the high attention regions for low-risk cases focused on tumor infiltrating lymphocytes, while the high attention regions for high-risk cases paid more attention to ulcerated regions and regions of densely packed tumor cells. For both high and low-risk cases, the low attention regions focused mainly on background blood. \textbf{b, h)} Local gene attributions for the corresponding low-risk (top) and high-risk (bottom) cases from a and g. \textbf{c, i)} Kaplan–Meier curves for omics-only (left, ``SNN"), histology-only (center, ``AMIL") and multimodal fusion (right, ``MMF"), showing improved separation using MMF.\textbf{d, j)} \rjc{Global gene attributions across patient cohorts according to unimodal interpretability (left, ``SNN"), and multimodal interpretability (right, ``MMF"), with gene RNA-Seq abundances receiving the most attribution. SNN and MMF were both able to identify immune-related and prognostic markers such as EGFR and PGRMC2 in HNSC, and DNAH8 and PRKCD in SKCM. MMF additionally attributes to other immune-related / prognostic genes such as C1ORF21, CSMD1, NRAS, and PRKCD.} \textbf{e, h)} High attention patches from low-risk (top) and high-risk (bottom) cases with corresponding cell labels. \textbf{f, i)} Quantification of cell types in high-attention patches for each disease overall. }
\end{figure*}

To obtain an understanding of how morphological features were used to explain cancer prognosis, we assessed high attention regions of WSIs in the top 25\% (high risk group) and bottom 25\% (low risk group) of predicted patient risks for each cancer type, which reflects favorable and poor cancer prognosis respectively. In addition to visual inspection from two pathologists, we simultaneously segmented and classified cell type identities across the top 1\% of high attention regions in our WSIs. \textbf{Fig. S13} displays the cell fraction distributions of each cell type across the two risk groups for all 14 cancer types, as well as exemplar high attention patches and cell identity instance segmentation results. Across all cancer types, we generally observed that high attention regions in low risk patients corresponded with greater immune cell presence and lower tumor grade than that of high risk patients, with 9/14 cancer types demonstrating statistically significant differences in lymphocyte cell fractions in high attention regions (\textbf{Fig. 3, Fig. 4, Fig. S1-10}). Furthermore, we also observed that high attention regions in high risk patients corresponded with increased tumor presence, higher tumor grade and tumor invasion in certain cancer types, with 7/14 cancer types demonstrating statistically significance differences in tumor cell fractions. These differences in cell fraction distributions suggests that attention-based interpretability uses lymphocyte and tumor cell presence as prognostic information to explain favorable and poor cancer prognoses respectively. \textbf{Fig. 4f} and \textbf{Fig. 4i} show clear differences in cell fraction distributions in head and neck squamous cell carcinoma (HNSC) and skin cutaneous melanoma (SKCM) in comparing tumor cell fraction (HNSC P-Value: $1.11 \times 10^{-9}$, SKCM P-Value: $6.26 \times 10^{-3}$, t-test) and lymphocyte cell fraction (HNSC P-Value: $4.60 \times 10^{-29}$, SKCM P-Value: $1.13 \times 10^{-10}$, t-test). \textbf{Fig. 4e} and \textbf{Fig. 4k} show exemplar high attention regions in low and high risk respectively, with attention-based interpretability identifying dense immune cell infiltrates (green) in low risk patients and nuclear pleomorphism and atypia in tumor cells (red) in high risk patients. Interestingly, increased fractional tumor cell content in high attention regions were not discovered in high risk patients for KIRC and KIRP. However, visual inspection of high attention regions in these cancer types revealed that tumor cells in low risk patients corresponded with lower tumor grade than that of high risk patients. \textbf{Fig. 3a} and \textbf{Fig. 3g} provide examples of attention heatmaps for low and high risk patients in KIRC and KIRP, in which high attention regions in high risk KIRC patients corresponded with central necrosis, and high attention regions in high risk KIRP correspond with tumor cells invading the renal capsule. \rjc{To understand how attention shifts when conditioning on molecular features in multimodal interpretability, we also had two trained pathologists use PORPOISE to assess unimodal and multimodal attention heatmaps. For certain cancer types such as BRCA and KIRC, attention in MMF shifted way from tumor-only regions and towards both stroma and tumor regions, which corroborates the prognostic relevance of stroma\cite{beck2011systematic, bejnordi2017deep}.} 

In parallel with assessing WSI interpretability, we also interrogated important model explanations in our molecular feature inputs. \textbf{Fig. S12} and \textbf{Fig. S13} display global gene attribution decision plots for mutation status, copy number variation, and RNA-Seq abundance across our 14 cancer types for SNN and MMF respectively. Since molecular features were interpreted using a gradient-based interpretability method, we are able to compute not only a numerical importance score, also the direction in which that feature contributes to low risk (negative, pointing left) or high risk (positive, pointing right). Across all cancer types, gradient-based interpretability was able to identify many well-known oncogenes and immune-related genes established in existing biomedical literature\cite{uhlen2017pathology}. In the combined glioblastoma and lower-grade glioma (GBMLGG) cohort, gradient-based interpretability identifies IDH1 mutation (P-Value: $3.77 \times 10^{-36}$, t-test) status as the most attributed gene feature, which has important functions in cellular metabolism, epigenetic regulation and DNA repair and defines the IDH1-wildtype astrocytoma, IDH1-mutant astrocytoma, and IDH-mutant oligodendroglioma molecular subtypes in GBMLGG\cite{louis2016who} (\textbf{Fig. S4d}). In addition, IDH1 mutation is associated with lower grade gliomas and thus favorable prognosis in comparison with IDH1-wildtype gliomas, which corroborates with the attribution direction of IDH1 mutation in the attribution decision plot, in which the distribution of IDH1 mutation attributions has only negative attribution values (low risk) (\textbf{Fig. S4d}). In GBMLGG, other gene features identified included ATRX mutation (P-Value, $2.408 \times 10^{-2}$, t-test), PTEN mutation (P-Value, $3.99 \times 10^{-44}$, t-test), PIK3CA mutation (P-Value, $6.19 \times 10^{-14}$, t-test), EGFR copy number variation (P-Value, $4.43 \times 10^{-29}$, t-test), MTAP copy number variation (P-Value, $9.00 \times 10^{-25}$, t-test), and SLC13A4 RNA-Seq expression (P-Value, $1.96 \times 10^{-60}$, t-test), which are implicated in functions such as controlling DNA repair, cell cycle, and angiogenesis\cite{brennan2013somatic} (\textbf{Fig. S4}). In addition IDH1 in GBMLGG, we also successfully identify several other key oncogenes in other cancer types such as KRAS in LUAD (P-Value, $5.13 \times 10^{-62}$, t-test), and VHL in KIRC (P-Value, $3.23 \times 10^{-9}$, t-test) (\textbf{Fig. 3d, Fig. S6d}). IN KIRC, in addition to known oncogenes such as VHL, ARID1A, TP53 and PTEN, immune-related genes such as CDKN2A copy number variation (P-Value, $1.16 \times 10^{-11}$, t-test) CDKN2B copy number variation (P-Value, $7.39 \times 10^{-13}$, t-test), ULK2 RNA-Seq expression (P-Value, $5.36 \times 10^{-28}$, t-test) and NFIX RNA-Seq (P-Value, $1.74 \times 10^{-24}$, t-test) were highly attributed, which corroborates with their roles in innate immunity and inflammatory cell signaling\cite{cancer2013comprehensive, uhlen2015tissue, chevrier2017immune, uhlen2017pathology, uhlen2019genome} (\textbf{Fig. 3d}). Across most cancer types, gene mutations that encode for extremely large proteins such as TTN, OBSCN, RYR3, and DNA5 were frequently found to be highly attributed. Though many of these genes are not explicitly cancer-associated and prognostic due to heterogeneity in the mutational processes of each cancer type, genomic instabilities in these large protein-coding domains may implicitly contribute to tumor mutational load\cite{lawrence2013mutational, rizvi2015mutational, shi2020exploring, oh2020spontaneous}. Attributions for all gene features for SNN and MMF can be found in the \textbf{Auxillary Supplement Materials.}


\section*{PORPOISE uses immune response to predict cancer prognosis}
\vspace{-4mm}

\begin{figure*}
\includegraphics[width=\textwidth]{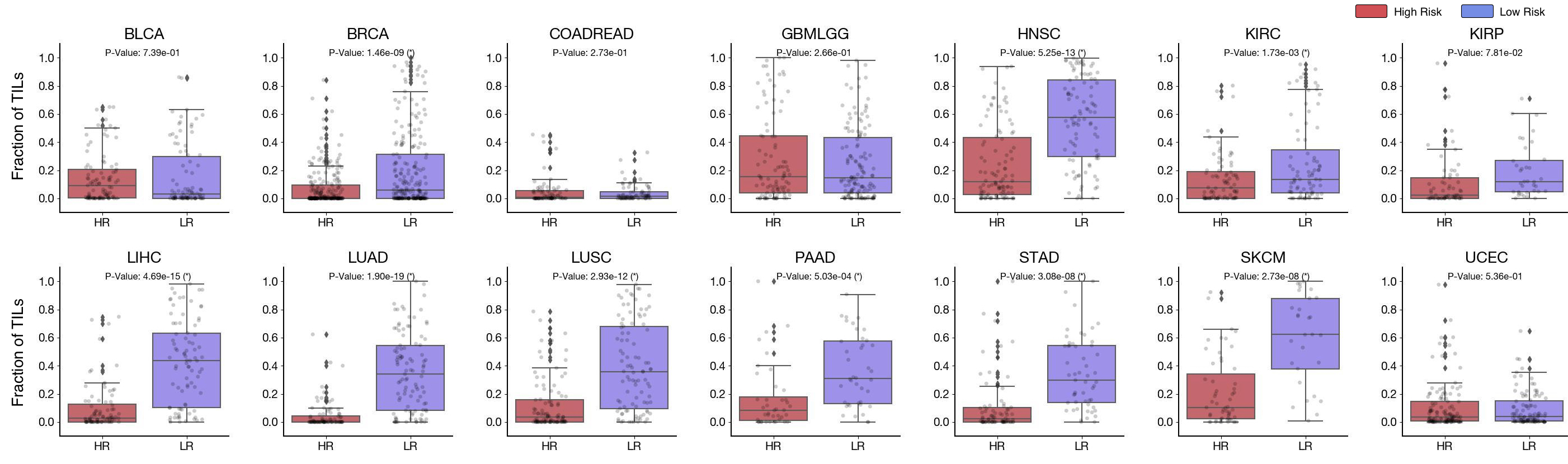}
\caption*{\textbf{Fig. 5: Tumor Infiltrating Lymphocyte Quantification in Patient Risk Groups} TIL quantification in high attention regions of predicted low and high risk patient cases across 14 cancer types. For each patient, the top 1\% of scored high attention regions ($512 \times 512$ $20\times$  image patches) were segmented and analyzed for tumor and immune cell presence. Image patches with high tumor-immune co-localization were indicated as positive for TIL presence (and negative otherwise). Across all patients, the fraction of high attention patches containing TIL presence was computed and visualized in the box plots. A two-sample t-test was computed for each cancer type to test the if the means of the TIL fraction distributions of low and high risk patients had a statistically significant difference (with * marked if P-Value $<$ 0.05).} 
\end{figure*}

Lastly, we used the interpretability of PORPOISE as a mechanism to test the hypothesis that tumor-infiltrating lymphocyte presence corroborates with favorable cancer prognosis. \textbf{Fig. 5} shows the fractional distribution of TILs in the top 1\% high attention regions for all 14 cancer types across the previously defined risk groups. In comparing TIL presence between low risk and high risk patients, we found that 9 out of 14 cancer types had a statistical significant increase in TIL presence amongst patients with predicted low risk, indicating that model attention was localized to more immune-hot regions. For cancer types in our dataset that have been FDA approved for immunotherapies, TIL presence was used as model explanations for favorable prognosis in BRCA (P-Value, $1.46 \times 10^{-9}$, t-test), HNSC (P-Value, $5.25 \times 10^{-13}$, t-test), KIRC (P-Value, $1.73 \times 10^{-3}$, t-test), LIHC (P-Value, $4.69 \times 10^{-15}$, t-test), LUAD (P-Value, $1.90 \times 10^{-19}$, t-test), LUSC (P-Value, $2.93 \times 10^{-12}$, t-test), PAAD (P-Value, $5.03 \times 10^{-4}$, t-test), STAD (P-Value, $3.08 \times 10^{-8}$, t-test), and SKCM (P-Value, $2.73 \times 10^{-8}$, t-test). This suggests that our trained models use morphological features for immune response as markers for predicting cancer prognosis, and supports a growing body of evidence that TILs have a prognostic role in many cancer types\cite{thorsson2018immune, saltz2018spatial, shaban2019novel, abduljabbar2020geospatial}. In breast cancer, Maley \textit{et al.} performed hotspot analysis on the co-localization of immune cancer cells in WSIs, and showed that immune-cancer co-localization was a significant predictive factor of long-term survival \cite{maley2015ecological}. In Oral Squamous Cell Carcinoma, Shaban \textit{et al.} proposed a  co-localization score for quantifying TIL density that showed similar findings\cite{shaban2019novel}. In lung cancer, AbdulJabbar \textit{et al.} proposed a deep learning framework for spatially profiling immune infilitration in H\&E and IHC WSIs, and similarly found that patients with more immune-cold regions had a strong correlation with disease-free survival\cite{abduljabbar2020geospatial}. Saltz \textit{et al.} performed a pan-cancer analysis on the spatial organization of TILs in the TCGA, and showed how different phenotypes of TIL infiltrates correlates with survival\cite{saltz2018spatial}. The distinction of these findings from our work, is that we demonstrate deep learning is able to discover TIL presence on its own as being a prognostic morphological feature for stratifying low and high risk patients.

\section*{\large{Discussion}}
\vspace{-4mm}
In this study, we present a method for interpretable, weakly-supervised, multimodal deep learning that integrates WSIs and molecular profile data for cancer prognosis. In comparison to previous works that selectively sample ROIs from WSIs for survival analysis, our approach is able to use the entire tissue and tumor microenvironment without any fine-grained annotation in WSIs, as well as multiple diagnostic slides of arbitrary sizes from a single patient sample during training and inference in an end-to-end framework. Following feature extraction from WSI and molecular profile inputs, our model learns a joint multimodal representation via the Kronecker Product that models pairwise features interactions, which is then used to estimate the hazard function in survival analysis. To demonstrate the effectiveness of our method, we trained and validated on 6,592 WSIs from 5,720 patients with paired molecular profile data across 14 cancer types, and compared our method with unimodal deep learning models as well as Cox models with clinical covariates, achieving the highest c-Index performance on 9 out of 14 cancer types in a one-versus-all comparison.

Our method is also the first to explore multimodal interpretability in explaining how features from WSIs and molecular features contribute towards risk. We developed PORPOISE, an interactive, freely available platform that directly yields both WSI and molecular feature explanations made by our model in our 14 cancer types. To validate that these model explanations are prognostic, we analyzed high attention morphological regions in WSIs, from which we discovered that tumor-infiltrating lymphocyte presence corroborates with favorable cancer prognosis on 9 out of 14 cancer types. This is one of many hypotheses that can be tested using PORPOISE. Our goal with PORPOISE is to begin making current black-box state-of-the-art methods in computational pathology, especially emerging multimodal methods, more transparent, explainable and usable for the broader biomedical research community. In making heatmaps and decision plots available for each cancer type, we hope that our tool would allow clinicians and researchers to devise their own hypotheses and investigate the discoveries explained using deep learning.

A limitation of our platform is that though PORPOISE can explain to us "what", it cannot always explain "why". For example, though TILs were used in a majority of our cancer types to distinguish low and high risk patients, post-hoc analyses still had to be done to validate that TIL presence was statistically significant between the two risk groups. One of the remarkable findings in our study was that in quantifying the contribution of WSIs in multimodal learning, WSIs on average accounted for only 10.0\% of input attributions. This is reflected in analyses on feature shift in WSIs, in which high attention shifted way from tumor regions to stroma, normal tissue and other morphological regions in the multimodal network, despite reporting the highest c-Index performance. We speculate this observation is a result of the intrinsic differences between WSIs and molecular profile data. In molecular features, the genotypic information from gene mutation, copy number variation, and RNA-Seq abundances have no spatial resolution are averaged across cells in the tumor biopsies, whereas phenotypic information such as normal tissue, tumor cells, and other morphological determinants are spatially represented in WSIs. As a result, when our multimodal algorithm is already conditioned with tumor-related features (\textit{e.g.} - TP53 mutation status, PTEN loss) in the molecular profile, it can attend to morphological regions with non-tumor information such as stroma to explain subtle differences in survival outcomes\cite{beck2011systematic}. Beyond characterizing human-identifiable phenotypes such as TILs, we hope that PORPOISE will be used by the research community towards identifying non-human-identifiable phenotypes that can be explained but not currently understood. 

As research advances in sequencing technologies such as single-cell RNA-Seq, mass cytometry and Spatial Transcriptomics continues to maturate and gain clinical penetrance, in combination with whole slide imaging, our approach to understanding molecular biology will become increasingly spatially-resolved and multimodal\cite{giesen2014highly, abdelmoula2016data, schapiro2017histocat, berglund2018spatial, jackson2020single}. In using bulk molecular profile data, our multimodal learning algorithm is considered a "late fusion" architecture, in which unimodal WSI and molecular features are fused towards the penultimate layers of the network\cite{baltruvsaitis2018multimodal}. Spatially-resolved genomics and transcriptomics data in combination with whole slide imaging has the potential to enable "early fusion" deep learning techniques that can integrate local histopathology image regions and molecular features with exact spatial correspondences, which could lead to more robust characterizations and spatial organization mappings of intratumoral heterogeneity, immune cell presence, and other morphological determinants.
\newpage

\vspace{-5mm}
\section*{\large{Methods}}
\begin{spacing}{1}
\vspace{-4mm}
\noindent\textbf{Dataset Description} \\
6,592 H\&E diagnostic WSIs with corresponding molecular and clinical data were collected from 5,720 patients across 14 cancer types from TCGA via the NIH Genomic Data Commons Data Portal. Sample inclusion criteria in dataset collection were defined by: 1) dataset size and balanced distribution of uncensored-to-censored patients in each TCGA project, and 2) availability of matching CNV, mutation, and RNA-Seq abundances for each WSI (WSI-CNV-MUT-RNA). To mitigate overfitting in modeling the survival distribution during survival analysis, projects with less than 150 patients (after WSI and molecular data alignment) and have poor uncensorship (less than 5\% uncensored patients) were excluded from the study. For two organs, brain and the gastrointestinal tract, cancer types from these organs were grouped together respectively, forming combined TCGA projects: GBMLGG (glioblastoma and lower-grade glioma) and COADREAD (colon and rectal adenocarcinoma). For inclusion of Skin Cutaneous Melanoma (TCGA-SKCM) and Uterine Corpus Endometrial Carcinoma (TCGA-UCEC) that have large data missingness, criteria for data alignment were relaxed to include samples with only matching WSI-MUT-RNA and WSI-CNV-MUT respectively. To decrease feature sparsity in molecular profile data, genes with greater than 5\% CNV or mutation frequency in each cancer study were used. To limit the number of the features from RNA-Seq, the top 2000 genes ordered by median absolute deviation were used. Molecular and clinical data were obtained from quality-controlled files from the cBioPortal. Summary tables of cohort characteristics, censorship statistics, and feature alignment can be found in \textbf{Table S1} and \textbf{Table S2}.

\noindent\textbf{WSI Processing} \\
For each WSI, automated segmentation of tissue was performed using the public CLAM\cite{lu2020data} repository for WSI analysis. Following segmentation, image patches of size $256 \times 256$ were extracted without overlap at the 20$\times$ equivalent pyramid level from all tissue regions identified. Subsequently, a ResNet50 model pretrained on Imagenet\cite{} is used as an encoder to convert each $256 \times 256$ patch into 1024-dimensional feature vector, via spatial average pooling after the 3rd residual block. To speed up this process, multiple GPUs were used to perform computation in parallel using a batch-size of 128 per GPU. 

\noindent\textbf{Deep Learning-based Survival Analysis for Integrating Whole Slide Images and Genomic Features} \\
PORPOISE (Pathology-Omics Research Platform for Integrated Survival Estimation) uses a high-throughput, interpretable, weakly-supervised, multimodal deep learning algorithm (MMF) designed for integrating whole slide images and molecular profile data in weakly-supervised learning tasks such as patient-level cancer prognosis via survival analysis. Given 1) diagnostic WSIs as pyramidal TIFF files and 2) processed genomic and transcriptomic features for a single patient, MMF learns to jointly represent these two heterogenous data modalities in an end-to-end deep learning algorithm. Though tasked for survival analysis, our algorithm is adaptable to any combination of modalities, and flexible for solving any learning tasks in computational pathology that have patient-level labels. Our algorithm consists of three components: 1) attention-based Multiple Instance Learning (AMIL) for processing WSIs, 2) Self-Normalizing Networks (SNN) for processing molecular profile data, and 3) a multimodal fusion layer (extended from Pathomic Fusion) for integrating WSIs and molecular profile data \cite{ilse2018attention, lu2020data, klambauer2017self, chen2020pathomic}.

\textit{AMIL.} To perform survival prediction from WSIs, we extend the attention-based multiple instance learning algorithm, which was originally proposed for weakly-supervised classification. Under the multiple instance learning framework, each gigapixel WSI is divided into smaller regions and viewed as a collection (known as a bag) of patches (known as instances) with a corresponding slide-level label used for training. Accordingly, following WSI processing, each WSI bag is represented by a  $M_i \times C$ matrix tensor, where $M_i$ is the number of patches (known as the bag size), which varies between slides, and C is the feature dimension and equals 1024 for the ResNet50 encoder we used. Since survival outcome information is available at the patient-level instead of for individual slides, we collectively treat all WSIs corresponding to a patient case as a single WSI bag during training and evaluation. Namely, for a patient case with $N$ WSIs with bag sizes $M_1, \cdots, M_N$ respectively, the WSI bag corresponding the patient is formed by concatenating all $N$ bags, and has dimensions $M \times 1024$, where $M = \sum_{i=1}^{N} M_i$.

The model can be described by three components, the projection layer $f_p$, the attention module $f_{attn}$, and the prediction layer $f_{pred}$. Incoming patch-level embeddings of each WSI bag, $\mathbf{H} \in \mathbbm{R}^{M \times 1024}$, are first mapped into a more compact, dataset-specific 512-dimensional feature space by the fully-connected layer $f_p$ with weights $\mathbf{W}_{proj} \in \mathbbm{R}^{512 \times 1024}$ and bias $\mathbf{b}_{bias} \in \mathbbm{R}^{512}$. For succinctness, from now on, we refer to layers using their weights only (the bias terms are implied). Subsequently, the attention module $f_{attn}$ learns to score each region for its perceived relevance to patient-level prognostic prediction. Regions with high attention scores contribute more to the patient-level feature representation relative to regions assigned low attention scores, when information across all regions in the patient's WSIs are aggregated, in an operation known as attention-pooling\cite{ilse2018attention}. Specifically, $f_{attn}$ consists of 3 fully-connected layers with weights $\mathbf{U}_{a} \in \mathbbm{R}^{256 \times 512}$, $\mathbf{V}_{a} \in \mathbbm{R}^{256 \times 512}$ and $\mathbf{W}_{a} \in \mathbbm{R}^{1 \times 256}$. Given a patch embedding $\mathbf{h}_{m} \in \mathbbm{R}^{512}$ (the $m^{th}$ row entry of $\mathbf{H}$), its attention score $a_{m}$ is computed by:
\begin{equation}
a_{m}=\frac{\exp \left\{\mathbf{W}_{a}\left(\tanh \left(\mathbf{V}_{a} \mathbf{h}_{m}^{\top}\right) \odot \operatorname{sigm}\left(\mathbf{U}_{a} \mathbf{h}_{m}^{\top}\right)\right)\right\}}{\sum_{m=1}^{M} \exp \left\{\mathbf{W}_{a} \left(\tanh \left(\mathbf{V}_{a} \mathbf{h}_{m}^{\top}\right) \odot \operatorname{sigm}\left(\mathbf{U}_{a}\mathbf{h}_{m}^{\top}\right)\right)\right\}}
\end{equation}
The attention-pooling operation then aggregates the patch-level feature representations into the patient representation ${\mathbf{h}_{patient}} \in \mathbbm{R}^{512}$ using computed attention scores as weight coefficients, where $\mathbf{A} \in \mathbbm{R}^M$ is the vector of attention scores:
\begin{equation} \label{clsweightedaverage}
{\mathbf{h}_{patient}}=\textbf{Attn-pool}(\mathbf{A}, \mathbf{H})=\sum_{m=1}^{M} a_{m} \mathbf{h}_{m}
\end{equation} 
The final patient-level prediction scores $\mathbf{s}$ are computed from the bag representation using the prediction layer $f_{pred}$ with weights $\mathbf{W}_{pred} \in \mathbbm{R}^{4 \times 512}$ and sigmoid activation: $\mathbf{s}=f_{pred}({\mathbf{h}_{bag}})$. This architectural choice and the negative-log-likelihood function for discrete-time survival modeling, are described in detail in a proceeding section. The last fully-connected layer is used to learn a representation $\mathbf{h}_{\text{WSI}} \in \mathbb{R}^{32 \times 1}$, which is then used as input to our multimodal fusion layer.

\textit{SNN.} To perform survival prediction from molecular features, we extended the Self-Normalizing Network (SNN) to receive high-dimensional, vector-concatenated molecular features of copy number variation, mutation, and RNA-Seq abundances. For learning scenarios that have hundreds to thousands of features with relatively few training samples, traditional Feedforward networks are prone to overfitting, as well as training instabilities from current deep learning regularization techniques such as stochastic gradient descent and Dropout. To employ more robust regularization techniques on high-dimensional low sample size genomics data, we adopted the normalizing activation and dropout layers from the SNN architecture: 1) scaled exponential linear units (SeLU) and 2) Alpha Dropout. In comparison with rectified linear unit (ReLU) activations common in current Feedforward networks, SeLU activations would drive the outputs of every layer towards zero mean and unit variance during layer propagation. The SeLU activation is defined as:
\begin{equation}
\operatorname{SeLU}(x) = \lambda\left\{\begin{array}{ll}x & \text { if } x>0 \\ \alpha e^{x}-\alpha & \text { if } x \leqslant 0\end{array}\right.
\end{equation} 
\noindent where $\alpha \approx 1.67, \lambda \approx 1.05$. To main normalization after Dropout, instead of setting the activation value to be $0$ with with probability $1-q$ for $0 < q \geq 1$ for a neuron in a given layer, the activation value is set to be $\lim _{x \rightarrow-\infty} \operatorname{SeLU}(x)=-\lambda \alpha=\alpha^{\prime}$, which ensures the self-normalization property with updated mean and variance $\mathbb{E}\left(x d+\alpha^{\prime}(1-d)\right)=q \mu+(1-q) \alpha^{\prime}$, $\operatorname{Var}\left(x d+\alpha^{\prime}(1-d)\right)=q\left((1-q)\left(\alpha^{\prime}-\mu\right)^{2}+\nu\right)$. The SNN architecture used for molecular feature input consists of 2 hidden layers of 256 neurons each, with SeLU activation and Alpha Dropout applied to every layer. The last fully-connected layer is used to learn a representation $\mathbf{h}_{\text{molecular}} \in \mathbb{R}^{32 \times 1}$, which is then used as input to our multimodal fusion layer.

\textit{Multimodal Fusion Layer.} Following the construction of unimodal feature representations from the AMIL and SNN subnetworks, we learn a multimodal feature representation using Kronecker Product Fusion, which would capture important interactions between these two modalities\cite{zadeh2017tensor, chen2020pathomic}. Our joint multimodal tensor is computed by the Kronecker product of $\mathbf{h}_{\text{WSI}}$ and $\mathbf{h}_{\text{molecular}}$, in which every neuron in $\mathbf{h}_{\text{molecular}}$ is multiplied by every other neuron in $\mathbf{h}_{\text{WSI}}$ to capture all bimodal interactions. To also preserve the unimodal features, we also append "1" to each unimodal feature presentation before fusion, which is shown the equation below:
\begin{equation}
    \mathbf{h}_{\text {fusion }}=\left[\begin{array}{l}\mathbf{h}_{\text{WSI}} \\ 1\end{array}\right] \otimes\left[\begin{array}{l}\mathbf{h}_{\text{molecular}} \\ 1\end{array}\right]
\end{equation}
where $\otimes$ is the Kronecker product, and $\mathbf{h}_{\text{fusion}} \in \mathbb{R}^{33 \times 33}$ is a differentiable multimodal tensor that models all unimodal and biomodal interaction with $O(1)$ computation. To decrease impact of noise unimodal features and to reduce feature collinearity between the WSI and molecular feature modalities, we used a gating-based attention mechanism that additionally controls the expressiveness of each modality:
\begin{equation}
\begin{aligned}
 \mathbf{h}_{i, \text{gated}} &= \mathbf{z}_i * \mathbf{h}_i, \forall \textbf{h}_i \in \{\mathbf{h}_\text{WSI}, \mathbf{h}_\text{molecular}\}\\ \text{where,  }
    \mathbf{h}_i &= \text{ReLU}(W_i \cdot \mathbf{h}_i)\\
    \mathbf{z}_i &= \sigma (W_{j} \cdot [\mathbf{h}_\text{WSI}, \mathbf{h}_\text{molecular}])
\end{aligned}
\end{equation}

For a modality $i$ with learned unimodal features $\textbf{h}_i$, we learn a weight matrix $W_{j}$ that would score the relative importance of each feature in modality $i$. After performing Softmax, $\mathbf{z}_i$ can be interpreted as an attention score of how $\mathbf{h}_\text{WSI}$ and $\mathbf{h}_\text{molecular}$ attends to each feature in $\textbf{h}_i$. We obtain the gated representation $\mathbf{h}_{i, \text{gated}}$ in taking the element-wise product of the original unimodal features $\textbf{h}_i$ and attention scores $\mathbf{z}_i$. Following gating-based feature attention and Kronecker product fusion, we propagate our multimodal tensor through two hidden layers of size 256, which is then subsequently supervised using a cross entropy-based loss function for survival analysis.

\textit{Survival Loss Function.} To perform survival prediction from right-censored, patient-level survival data, we first partition the continuous time scale of overall patient survival time in months, $T_{cont}$ into 4 non-overlapping bins: $[t_0, t_1), [t_1, t_2), [t_2, t_3), [t_3, t_4)$, where $t_0 = 0$, $t_4 = \infty$ and $t_1, t_2, t_3$ define the quartiles of event times for uncensored patients. Subsequently, for each patient entry in the dataset, indexed by $j$ with corresponding follow-up time $T_{j, cont} \in [0, \infty)$, we define the discretized event time $T_{j}$ as the index of the bin interval that contains $T_{j, cont}$: 
\begin{equation} 
T_{j} = r \text{ iff } T_{j, cont} \in [t_r, t_{r+1})
\end{equation}
To avoid confusion, we refer to the discretized ground truth label of the $j^{th}$ patient as $Y_j$. For a given patient with bag-level representation ${\mathbf{h}_{bag}}_{j}$, the prediction layer $f_{pred}$ with weights $\mathbf{W}_{pred} \in \mathbbm{R}^{4 \times 512}$ models the hazard function defined as:  
\begin{equation}
f_{hazard}(r \mid {\mathbf{h}_{bag}}_{j}) = P(T_j=r \mid T_j\geq r, {\mathbf{h}_{bag}}_{j})
\end{equation}
which relates to the survival function through:
\begin{equation}
\begin{split}
f_{surv}(r \mid {\mathbf{h}_{bag}}_{j}) & = P(T_j > r \mid  {\mathbf{h}_{bag}}_{j})\\
& = \prod_{u=1}^{r} (1 - f_{hazard}(u \mid {\mathbf{h}_{bag}}_{j}))
\end{split}
\end{equation}
To optimize the model parameters, we use the log likelihood function for a discrete survival model\cite{zadeh2020bias}, which given the binary censorship status $c_j$, can be expressed as 
\begin{equation}
\begin{aligned}
L = - l & = - c_{j} \cdot \log \left(f_{surv}\left(Y_j \mid {\mathbf{h}_{bag}}_{j} \right)\right)\\ & - \left(1-c_j\right) \cdot \log \left(f_{surv}\left(Y_j-1 \mid {\mathbf{h}_{bag}}_{j} \right)\right)\\
& -\left(1-c_j\right) \cdot \log \left(f_{hazard}\left(Y_j \mid {\mathbf{h}_{bag}}_{j} \right)\right)
\end{aligned}
\end{equation}
In this formulation, we use $c_j = 1$ for patients who have lived past the end of the follow-up period and $c_j = 0$ in the event that the patient passed away precisely at time $T_{j, cont}$. During training, the contribution of uncensored patient cases can be emphasized by minimizing a weighted sum of $L$ and $L_{uncensored}$ 
\begin{equation}
L_{surv} = (1 - \beta) \cdot L + \beta \cdot L_{uncensored} 
\end{equation}
The second term of the loss function corresponding uncensored patients, is defined by:
\begin{equation}
\begin{aligned}
L_{uncensored} = & - \left(1-c_j\right) \cdot \log \left(f_{surv}\left(Y_j-1 \mid {\mathbf{h}_{bag}}_{j} \right)\right) \\
& -\left(1-c_j\right) \cdot \log \left(f_{hazard}\left(Y_j \mid {\mathbf{h}_{bag}}_{j} \right)\right)
\end{aligned}
\end{equation}\

\textit{Training Details.} Across all cancer types, MMF is trained end-to-end with AMIL subnetwork, SNN subnetwork and multimodal fusion layer, using Adam optimization with a learning rate of $2\times 10^{-4}$, $b_1$ coefficient of 0.9, $b_2$ coefficient of 0.999, $\mathcal{L}_2$ weight decay of $1\times 10^{-5}$, and $\mathcal{L}_1$ weight decay of $1\times 10^{-4}$ for $20$ epochs. Because WSIs across patient samples have varying image dimension sizes, we randomly sampled paired WSIs and molecular features with a mini-batch size of 1. In performing comparative analyses with unimodal networks, AMIL and SNN were also also trained independently using the same survival loss function and hyperparameters as MMF.

\noindent\textbf{Multimodal Interpretability and Visualization} \\
\indent\textit{Local WSI Interpretability.} For a given WSI, to perform visual interpretation of the relative importance of different tissue regions towards the patient-level prognostic prediction, we first compute attention scores for 256 $\times$ 256 patches (without overlap) from all tissue regions in the slide. We refer to the attention score distribution across all patches from all WSIs from the patient case as the reference distribution. For fine-grained attention heatmaps, attention scores for each WSI are recomputed by increasing the tiling overlap to up to 90\%. For visualization, the attention scores are converted to percentile scores between 0.0 (low attention) to 1.0 (high attention) using the initial reference distribution, and spatially registered onto the corresponding WSIs (scores from overlapping patches are averaged). The resulting heatmap, referred to as local WSI interpretability, is transformed to RGB values using a colormap and overlayed onto the original H\&E image with a transparency value of 0.5.

\indent\textit{Global WSI Interpretability.} For sets of WSIs belonging to different patient cohorts, we performed global WSI interpretability by quantifying and characterizing the morphological patterns in the highest-attended image patches from each WSI. Since WSIs have differing image dimensions, we extracted a proportional amount of high attention image patches (1\%) to the total image dimension. On average, each WSI contained 13,487 $512 \times 512$ $20\times$ images, with approximately 135 image patches used as high attention regions. These attention patches are analyzed using a HoverNet model pretrained for simultaneous cell instance segmentation and classification\cite{graham2019hover}. Cells are classified as either tumor cells (red), lymphocytes (green), connective tissue (blue), dead cells (yellow), or non-neoplastic epithelial cells (orange). For each of these cell types, we analyzed the cell type frequency across all counted cells in the highest-attended image patches in a given patient, then analyzed the cell fraction distribution across all patients in low risk and high risk patients, defined as patients below and above the 25\% and 75\% predicted risk percentiles respectively.

\indent\textit{Tumor-Infiltrating Lymphocyte Detection.} To detect Tumor-Infiltrating Lymphocyte (TIL) presence in image patches, similar to other work, we defined TIL presence as the co-localization of tumor and immune cells which reflects intratumoral TIL response\cite{maley2015ecological, shaban2019novel}. Following cell instance segmentation and classification of tumor and immune cells in the highest-attended $512 \times 512$ $20\times$ image patches, we defined a heuristic which classified an image patch as positive for TIL presence with high tumor-immune cell co-localization (patch containing more than 20 counted cells, and more than 10 detected lymphocytes and 5 detected tumor cells). Similar to computing cell fraction distributions, for the highest-attended image patches in a given patient, we computed the fraction of TIL positive image patches, and plotted its distribution in low and high risk patients.

\textit{Local and Global SNN Interpretability.} For a given set of molecular features $x$ belonging to a patient sample, to characterize feature importance magnitude and direction of impact, we used Integrated Gradients (IG), a gradient-based feature attribution method that attributes the prediction of deep networks to their inputs \cite{sundararajan2017axiomatic}. IG satisfies two axioms for interpretability: 1) Sensitivity, in which for every desired input $x$ and baseline $x_i$ that differ in one feature but have different predictions, the differing feature should be given a non-zero attribution, and 2) Implementation Invariance, which states that two networks are functionally equivalent if their outputs are equal for all inputs. For a given input $x$, IG calculates the gradients of $x$ across different scales against a zero-scaled baseline $x_i$, which then uses the Gauss-Legendre quadrature to approximate the integral of gradients.

\begin{equation}
\text{IG}_{i}(x)::=\left(x_{i}-x_{i}^{\prime}\right) \times \int_{\alpha=0}^{1} \frac{\partial F\left(x^{\prime}+\alpha \times\left(x-x^{\prime}\right)\right)}{\partial x_{i}} d \alpha
\end{equation}

Using IG, for each molecular feature in input $x$ belonging to a patient sample, we compute feature attribution values, which corresponds to the magnitude of how much varying that feature in $x$ will change the output. Features that have no impact on the output would have zero attribution, whereas features that affect the output would have larger magnitude (interpreted as feature importance). In the context of regression tasks such as survival analysis, features that are positive attribution contribute towards increasing the output value (high risk), whereas negative attribution corresponds with decreasing the output value (low risk). For individual samples, we can use IG to understand how molecular features contribute towards the model risk prediction, which we can visualize as bar plots (termed local interpretability), where the x-axis corresponds with attribution value, the y-axis ranks features in order of absolute attribution magnitude (in descending order), and color corresponds with feature value. For binary data such as mutation status, bar colors are either colored blue (feature value '0', wild-type) or red (feature value '1', or mutation). For categorical and continuous data such as copy number variation and RNA-Seq abundance, bar colors are visualized using heatmap colors, where blue is low feature value (copy number loss / low RNA-Seq abundance) and red is high feature value (copy number gain / high RNA-Seq abundance). For large cohorts of patients from a cancer type, we can visualize the distribution of feature attributions across all patients (termed global interpretability), where each dot represents the attribution and feature value of an individual feature of an individual patient sample. Plots and terminology for local and global interpretability were derived from decision plots in Shapley Additive Explanation-based methods\cite{lundberg2020local}. 

\textit{Measuring WSI Contribution in Model Prediction.} To measure the contribution of WSIs in model predictions, for each patient sample, we compute the attributions for each modality at the penultimate hidden layer before multimodal fusion (last layer of the SNN and AMIL subnetworks). Then, we normalize the sum of absolute attribution values for each modality, to estimate percentage that each modality contributes towards the model prediction\cite{kokhlikyan2020captum}. 

\noindent\textbf{Evaluation Details and Statistical Analysis.} The predicted risk scores for AMIL, SNN, and MMF across all cancer types were evaluated on the same validation splits in a 5-fold cross-validation. To plot the Kaplan-Meier curves, we pooled out-of-sample risk predictions from the validation folds and plotted them against their survival time. For significance testing of patient stratification in Kaplan-Meier analysis, we use the logrank test to measure if the difference of two survival distributions is statistically significant (P-Value $<$ 0.05) \cite{bland2004logrank}. Cross-validated c-Index performance is reported as the average c-Index over the  5-folds. To estimate 95\% confidence intervals in cross-validation, we performed non-parametric bootstrapping using 1000 replicates on the out-of-sample predictions in the validation folds\cite{ledell2015computationally, tsamardinos2018bootstrapping}. For assessing global morphological feature significance of individual cell type presence, two-sample t-tests were performed in evaluating the statistical significance of mean cell fraction distributions in the top 1\% of high attention regions of low and high risk patients (P-Value $<$ 0.05). For assessing global molecular feature significance of individual gene features, two-sample t-tests were performed in evaluating the statistical significance of median gene attributions of low and high gene feature values (below and above median gene feature value respectively). For all box plots, boxes indicate the 1st, median, and 3rd quartile values of the data distribution, and whiskers extend to data points within 1.5$\times$ the interquartile range.

\noindent\textbf{Computational Hardware and Software} \\
PORPOISE was built with the OpenSeaDragon API and is hosted on Google Cloud. Python (version 3.7.7) packages used by PORPOISE include PyTorch (version 1.3.0), Lifelines (version 0.24.6), NumPy (version 1.18.1), Pandas (version 1.1.3), PIL (version 7.0.0), and OpenSlide (version 1.1.1). All WSIs were processed on Intel Xeon multi-core CPUs (Central Processing Units) and a total of 16 2080 Ti GPUs (Graphics Processing Units) using our custom, publicly available CLAM\cite{lu2020data} whole slide processing pipeline. The multimodal fusion layer for integrating WSIs and molecular profiles was implemented using our custom, publicly available Pathomic Fusion\cite{chen2020pathomic} software implemented in Python. Deep learning models were trained with Nvidia softwares CUDA 11.0 and cuDNN 7.5. Integrated Gradients was implemented using Captum (version 0.2.0)\cite{kokhlikyan2020captum}. Cell instance segmentation and classification was implemented using the HoVerNet software\cite{graham2019hover}. Statistical analyses such as two-sampled t-tests and logrank tests used implementations from SciPy (1.4.1) and Lifelines (version 0.24.6) respectively. Plotting and visualization packages were generated using Seaborn (0.9.0), Matplotlib (version 3.1.1), and Shap (0.35.0). 
\end{spacing}
\section*{\large{References}}
\vspace{2mm}
\begin{spacing}{0.9}
\bibliographystyle{naturemag}
\bibliography{sample}
\end{spacing}

\vspace{-4mm}
\noindent\textbf{Funding:} This work was supported in part by internal funds from BWH Pathology, Google Cloud Research Grant and Nvidia GPU Grant Program and NIGMS R35GM138216 (F.M.). R.J.C. was additionally supported by the NSF Graduate Fellowship. M.S. was additionally supported by the NIH Biomedical Informatics and Data Science Research Training Program, grant number: NLM T15LM007092. The content is solely the responsibility of the authors and does not reflect the official views of the National Institutes of Health, National Institute of General Medical Sciences or the National Library of Medicine. \\
\noindent\textbf{Authors contributions}: R.J.C. and F.M. conceived the study and designed the experiments. R.J.C. and M.Y.L. performed the experimental analysis. R.J.C. M.Y.L. Z.N. developed data visualization tools. R.J.C. M.Y.L. D.W. T.Y.C. M.S. J.L. M.W. B.J. M.S. F.M. analyzed the results. R.J.C. F.M. prepared the manuscript. F.M. supervised the research.\\
\noindent\textbf{Competing Interests}: The authors declare that they have no competing financial interests. \\
\noindent\textbf{Data and materials availability:} All diagnostic whole slide images and their corresponding molecular and clinical data are publicly accessible through the NIH Genomic Data Commons Data Portal. All code was implemented in Python, using PyTorch as the primary deep learning package. All code and scripts to reproduce the experiments of this paper are available at: \href{ https://github.com/mahmoodlab/PORPOISE}{ https://github.com/mahmoodlab/PORPOISE}. \\
\noindent\textbf{Ethics Oversight:} The study was approved by the Mass General Brigham (MGB) IRB office under protocol 2020P000233.
\section*{\large{List of Supplementary Materials}}
\vspace{-4mm}
Fig.  S1 - S14 (See below) \\
Table S1 - S5  (See below)\\
Auxiliary Supplementary Materials (See excel sheets) 
\end{spacing}

\begin{figure*}
\vspace{-9mm}
\includegraphics[width=\textwidth]{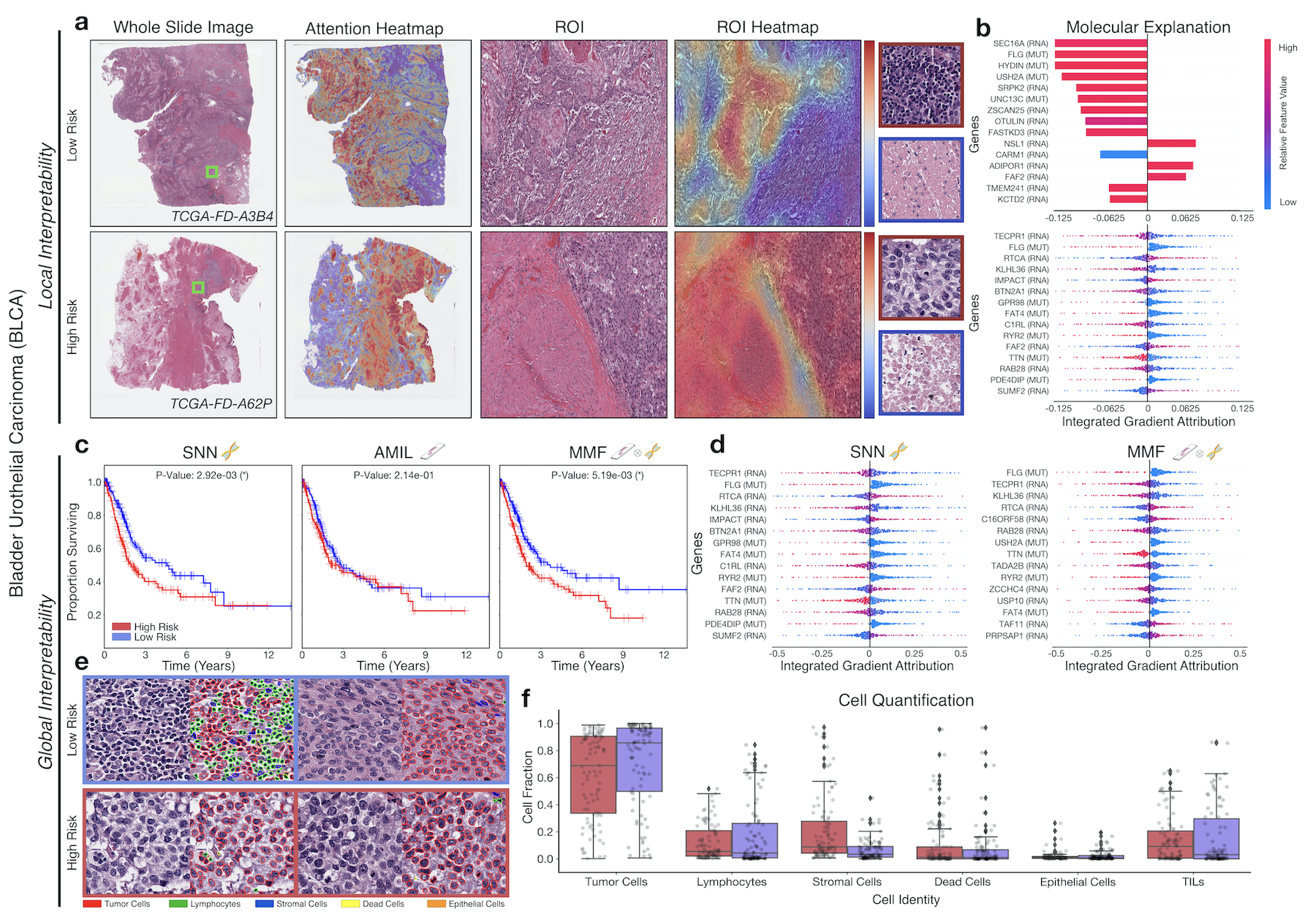}
\caption*{\textbf{Fig. S1: Quantitative performance, local model explanation, and global interpretability analyses of PORPOISE on BLCA.} \textbf{a.} WSIs, associated attention heatmaps, ROIs, ROI heatmaps, and selected high-attention patches from example low-risk (top) and high-risk (bottom) cases. In BLCA, high attention for low-risk cases tends to focus on aggregates of lymphocytes, thick tumor papillae, and muscularis, while in high-risk cases, high attention focuses on sheet-like and solid tumor growth, muscularis, and areas of necrosis. \textbf{b.} Local gene attributions for the corresponding low-risk (top) and high-risk (bottom) cases. \textbf{c.} Kaplan–Meier curves for omics-only (left, "SNN"), histology-only (center, "AMIL") and multimodal fusion (right, "MMF"), showing improved patient stratification over AMIL and long-surviving patients in SNN. \textbf{d.} Global gene attributions across patient cohorts according to unimodal interpretability (left, "SNN"), and multimodal interpretability (right, "MMF"). \textbf{e.} High attention patches from low-risk (top) and high-risk (bottom) cases with corresponding cell labels. \textbf{f.} Quantification of cell types in high-attention patches for all cases of BLCA with high risk in red and low risk in blue, showing high tumor cell abundance in both risk groups, with increased stromal cell presence in high risk groups.}
\end{figure*}
\clearpage

\begin{figure*}
\vspace{-9mm}
\includegraphics[width=\textwidth]{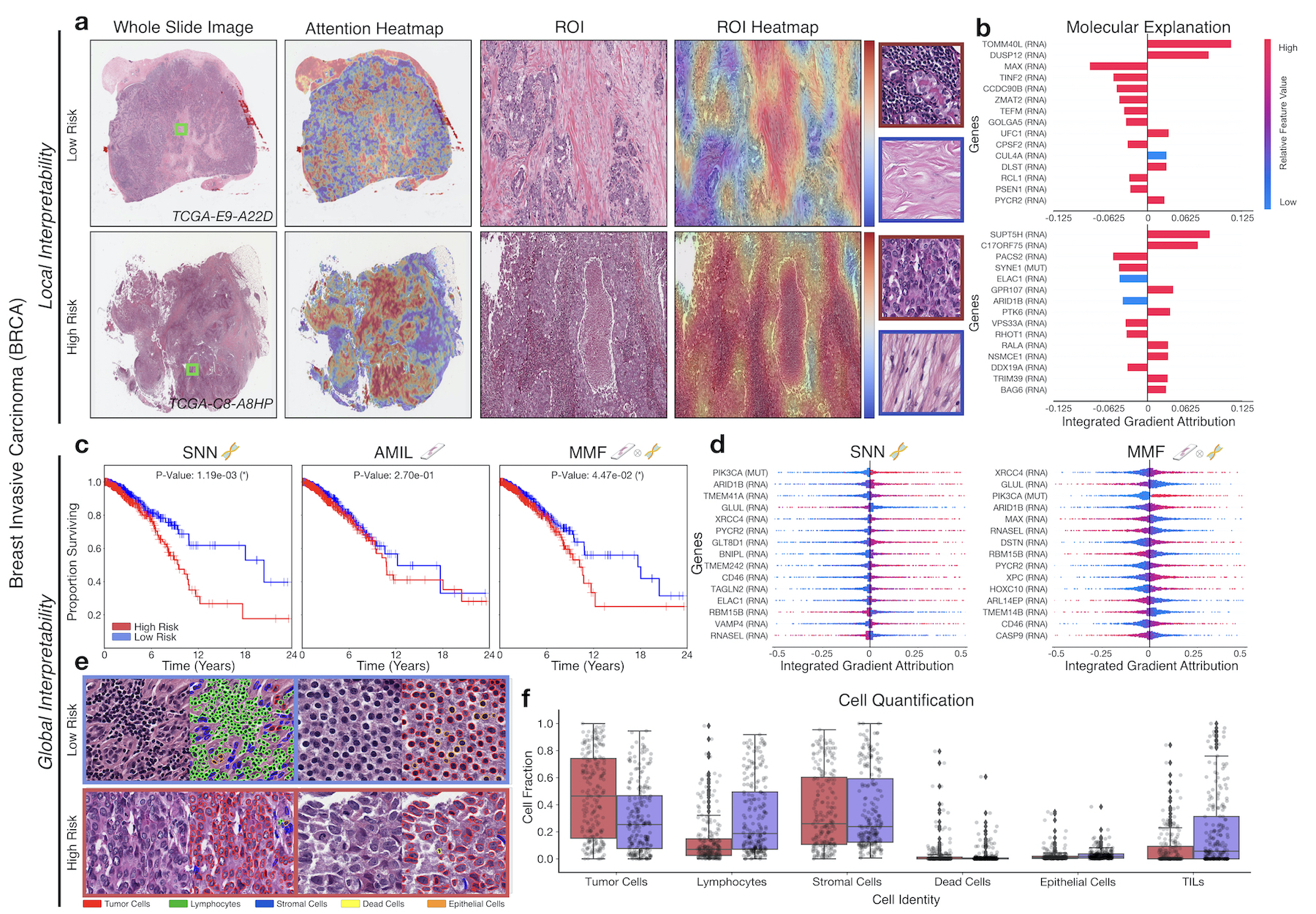}
\caption*{\textbf{Fig. S2: Quantitative performance, local model explanation, and global interpretability analyses of PORPOISE on BRCA.} \textbf{a.} WSIs, associated attention heatmaps, ROIs, ROI heatmaps, and selected high-attention patches from example low-risk (top) and high-risk (bottom) cases. In BRCA, high attention for low-risk cases tends to focus on collagenous stroma and aggregates of lymphocytes, while in high-risk cases, high attention focuses on areas of tumor cells with increased mitotic activity, nuclear pleomorphism, and necrosis. \textbf{b.} Local gene attributions for the corresponding low-risk (top) and high-risk (bottom) cases. \textbf{c.} Kaplan–Meier curves for omics-only (left, "SNN"), histology-only (center, "AMIL") and multimodal fusion (right, "MMF"), showing improved patient stratification over AMIL and long-surviving patients in SNN. \textbf{d.} Global gene attributions across patient cohorts according to unimodal interpretability (left, "SNN"), and multimodal interpretability (right, "MMF"). \textbf{e.} High attention patches from low-risk (top) and high-risk (bottom) cases with corresponding cell labels. \textbf{f.} Quantification of cell types in high-attention patches for each disease overall, showing decreased tumor cell abundance and increased lymphocyte and TIL presence in low risk groups.}
\end{figure*}
\clearpage

\begin{figure*}
\vspace{-9mm}
\includegraphics[width=\textwidth]{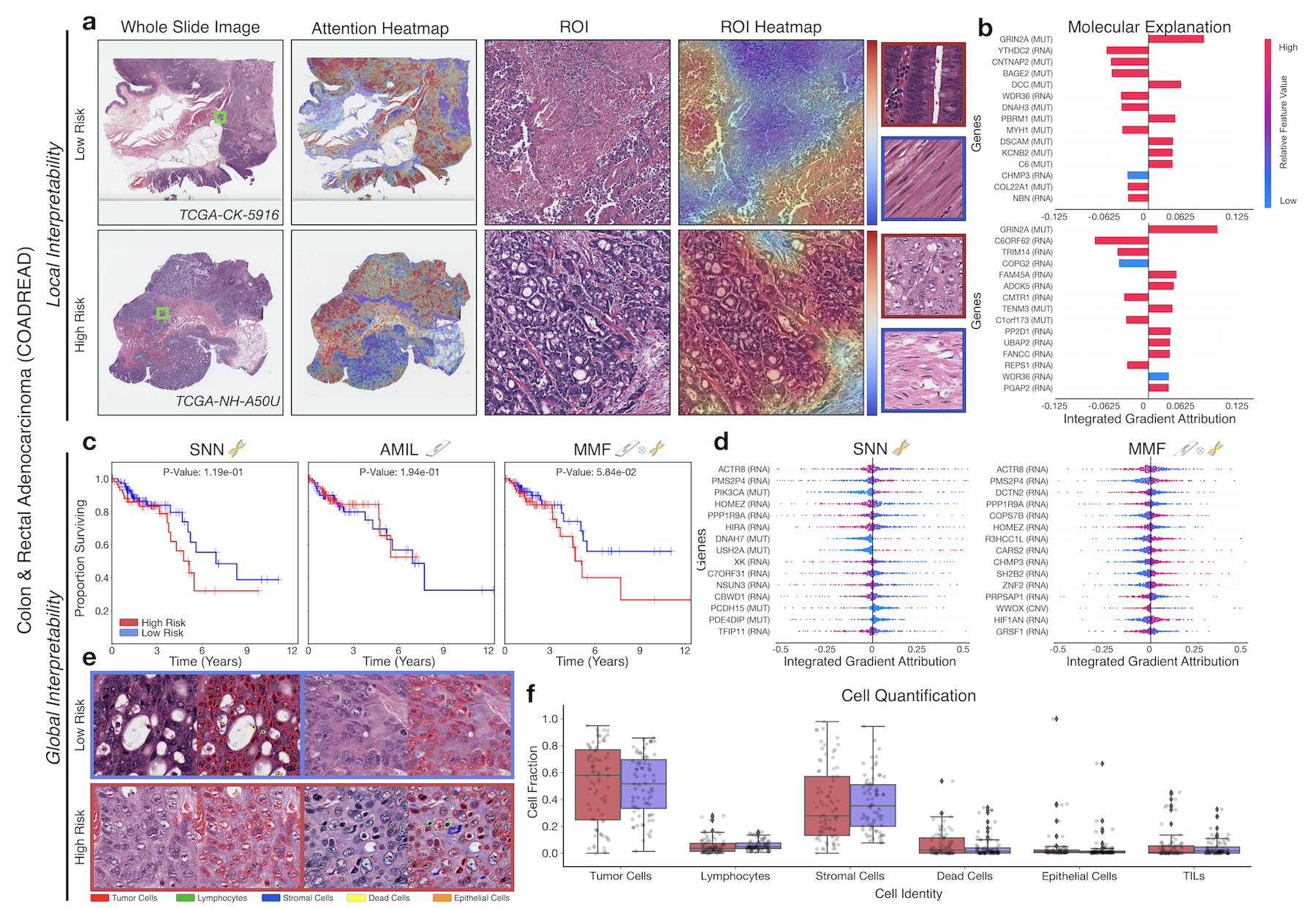}
\caption*{\textbf{Fig. S3: Quantitative performance, local model explanation, and global interpretability analyses of PORPOISE on COADREAD.} \textbf{a.} WSIs, associated attention heatmaps, ROIs, ROI heatmaps, and selected high-attention patches from example low-risk (top) and high-risk (bottom) cases. In COADREAD, high attention for low-risk cases tends to focus on muscularis, solid tumor growth and small nests of tumor cells, while in high-risk cases, high attention focuses on tumor cells invading the submucosa into the musclaris. \textbf{b.} Local gene attributions for the corresponding low-risk (top) and high-risk (bottom) cases. \textbf{c.} Kaplan–Meier curves for omics-only (left, "SNN"), histology-only (center, "AMIL") and multimodal fusion (right, "MMF"), showing improved patient stratification over AMIL and long-surviving patients in SNN. \textbf{d.} Global gene attributions across patient cohorts according to unimodal interpretability (left, "SNN"), and multimodal interpretability (right, "MMF"). \textbf{e.} High attention patches from low-risk (top) and high-risk (bottom) cases with corresponding cell labels. \textbf{f.} Quantification of cell types in high-attention patches for each disease overall, showing similar admixtures of tumor cells, lymphocytes, and stromal cells across both risk groups.}
\end{figure*}
\clearpage

\begin{figure*}
\vspace{-9mm}
\includegraphics[width=\textwidth]{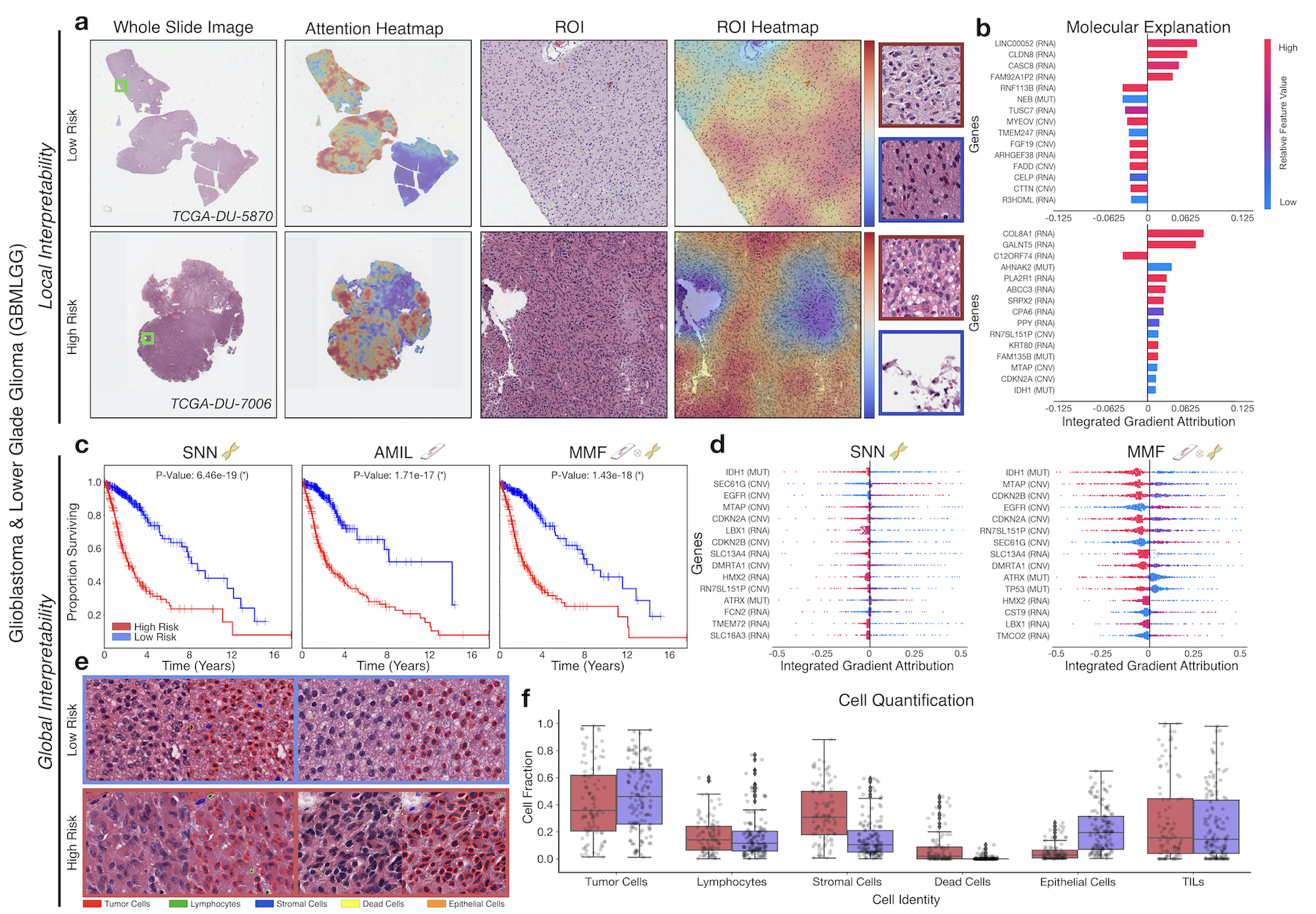}
\caption*{\textbf{Fig. S4: Quantitative performance, local model explanation, and global interpretability analyses of PORPOISE on GBMLGG.} \textbf{a.} WSIs, associated attention heatmaps, ROIs, ROI heatmaps, and selected high-attention patches from example low-risk (top) and high-risk (bottom) cases. In GBMLGG, high attention for low-risk cases tends to focus on dense regions of tumor cells, while in high-risk cases, high attention focuses on both dense regions of tumor cells and areas of vascular proliferation. \textbf{b.} Local gene attributions for the corresponding low-risk (top) and high-risk (bottom) cases. \textbf{c.} Kaplan–Meier curves for omics-only (left, "SNN"), histology-only (center, "AMIL") and multimodal fusion (right, "MMF"), with statistically significant patient stratification between low and high risk groups across all models. \textbf{d.} Global gene attributions across patient cohorts according to unimodal interpretability (left, "SNN"), and multimodal interpretability (right, "MMF"). \textbf{e.} High attention patches from low-risk (top) and high-risk (bottom) cases with corresponding cell labels. \textbf{f.} Quantification of cell types in high-attention patches for each disease overall, showing similar admixtures of tumor cells, lymphocytes, and stromal cells across both risk groups.}
\end{figure*}
\clearpage

\begin{figure*}
\vspace{-9mm}
\includegraphics[width=\textwidth]{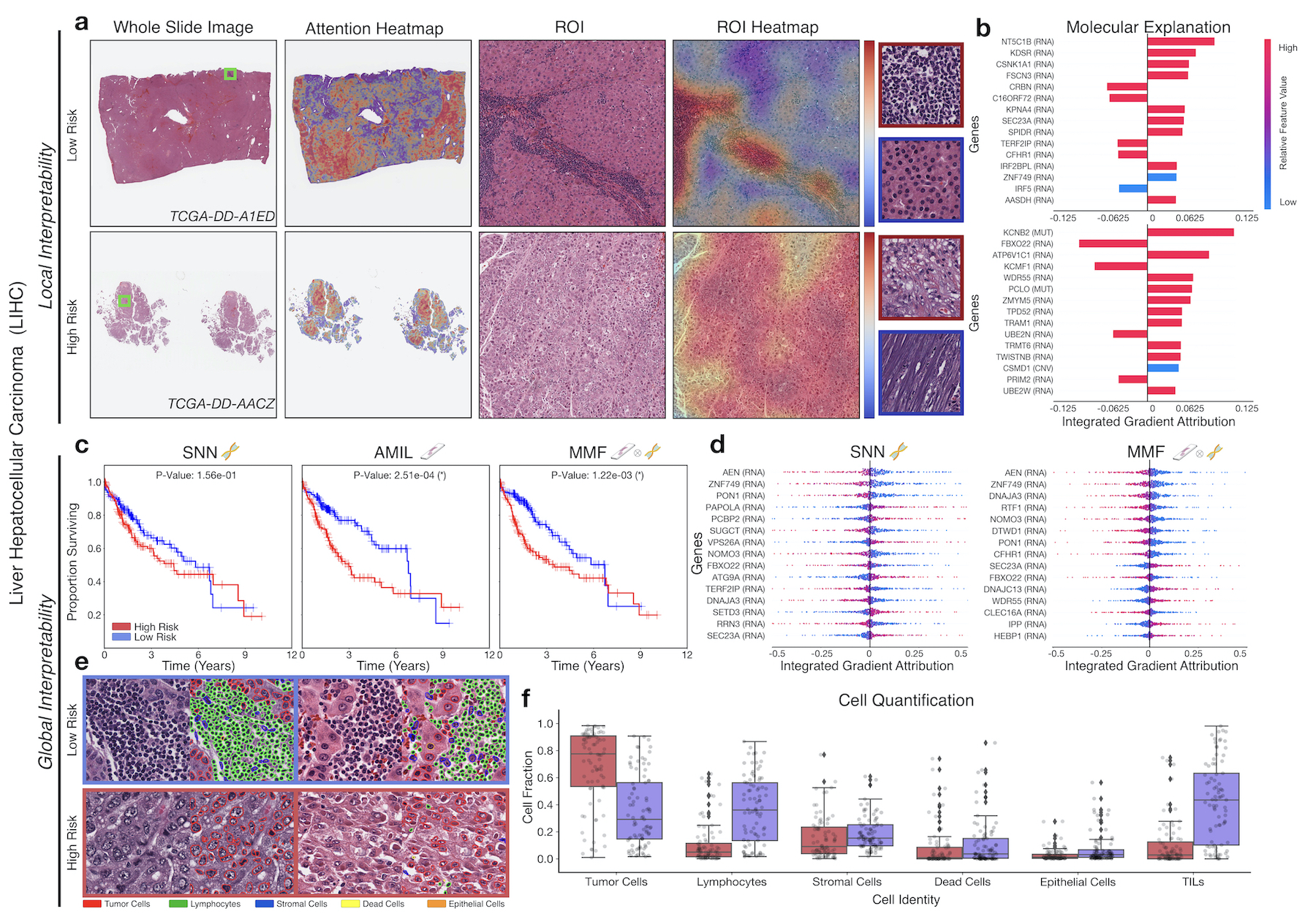}
\caption*{\textbf{Fig. S5: Quantitative performance, local model explanation, and global interpretability analyses of PORPOISE on LIHC.} \textbf{a.} WSIs, associated attention heatmaps, ROIs, ROI heatmaps, and selected high-attention patches from example low-risk (top) and high-risk (bottom) cases. In LIHC, high attention for low-risk cases tends to focus on dense regions of lymphocytes, while in high-risk cases, high attention focuses on areas with high tumor-grade morphology, such as increased nuclear pleomorphism. \textbf{b.} Local gene attributions for the corresponding low-risk (top) and high-risk (bottom) cases. \textbf{c.} Kaplan–Meier curves for omics-only (left, "SNN"), histology-only (center, "AMIL") and multimodal fusion (right, "MMF"), showing poor stratification with SNN and better stratification in AMIL and MMF. \textbf{d.} Global gene attributions across patient cohorts according to unimodal interpretability (left, "SNN"), and multimodal interpretability (right, "MMF"). \textbf{e.} High attention patches from low-risk (top) and high-risk (bottom) cases with corresponding cell labels. \textbf{f.} Quantification of cell types in high-attention patches for each disease overall, showing increased tumor cell presence in high risk patients and increased lymphocyte and TIL presence in low risk patients.}
\end{figure*}
\clearpage

\begin{figure*}
\vspace{-9mm}
\includegraphics[width=\textwidth]{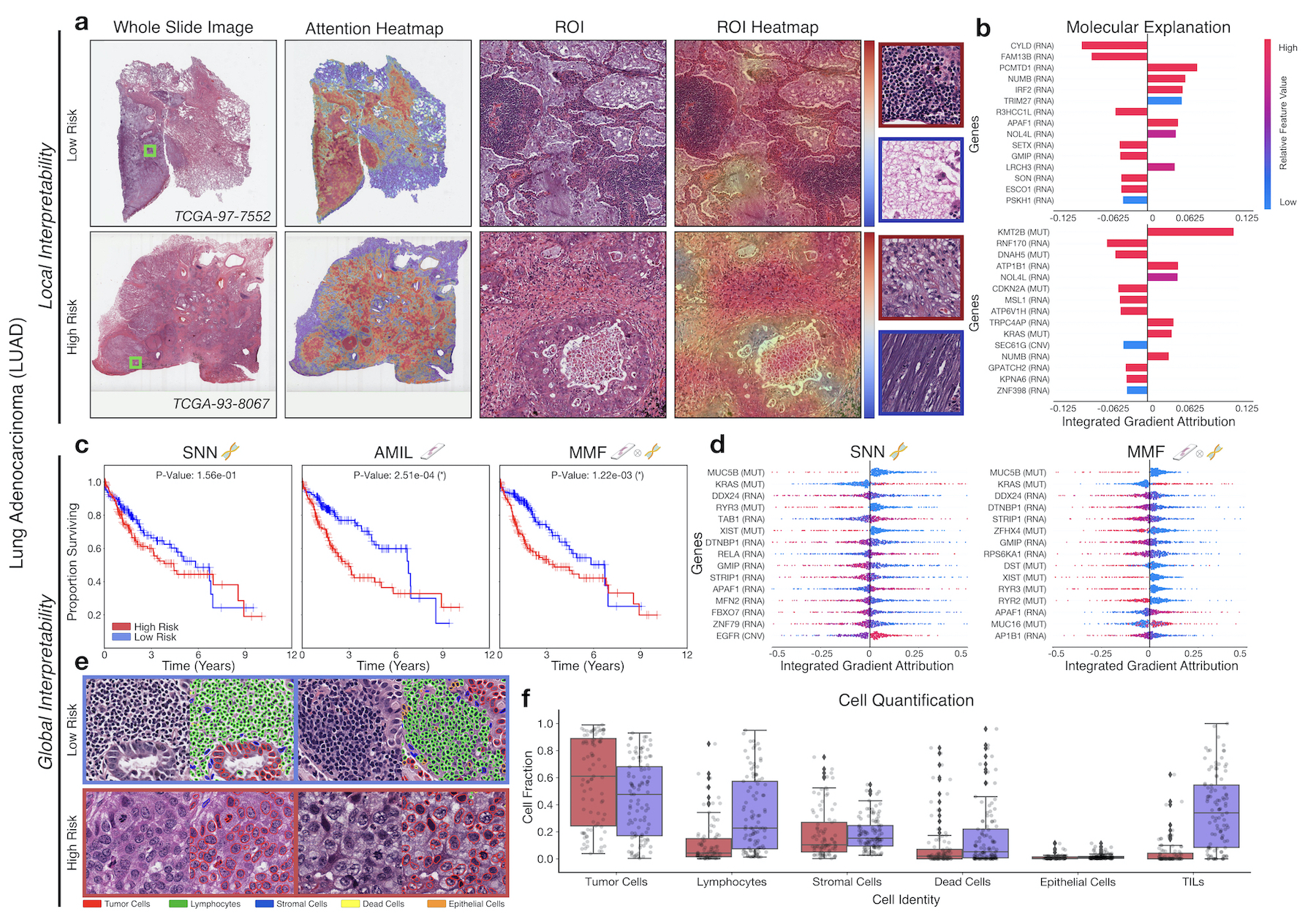}
\caption*{\textbf{Fig. S6: Quantitative performance, local model explanation, and global interpretability analyses of PORPOISE on LUAD.} \textbf{a.} WSIs, associated attention heatmaps, ROIs, ROI heatmaps, and selected high-attention patches from example low-risk (top) and high-risk (bottom) cases. In LUAD, high attention for low-risk cases tends to focus on regions with dense inflammatory infiltrate, predominantly comprised of lymphocytes, and regions of mucin deposition, while in high-risk cases, high attention focuses on tumor cells with increased nuclear pleomorphism, areas of necrosis, and tumor-associated dense fibrous stroma. \textbf{b.} Local gene attributions for the corresponding low-risk (top) and high-risk (bottom) cases. \textbf{c.} Kaplan–Meier curves for omics-only (left, "SNN"), histology-only (center, "AMIL") and multimodal fusion (right, "MMF"), showing poor stratification with SNN and better stratification in AMIL and MMF \textbf{d.} Global gene attributions across patient cohorts according to unimodal interpretability (left, "SNN"), and multimodal interpretability (right, "MMF"). \textbf{e.} High attention patches from low-risk (top) and high-risk (bottom) cases with corresponding cell labels. \textbf{f.} Quantification of cell types in high-attention patches for each disease overall, showing increased lymphocyte and TIL presence in low risk patients.}
\end{figure*}
\clearpage

\begin{figure*}
\vspace{-9mm}
\includegraphics[width=\textwidth]{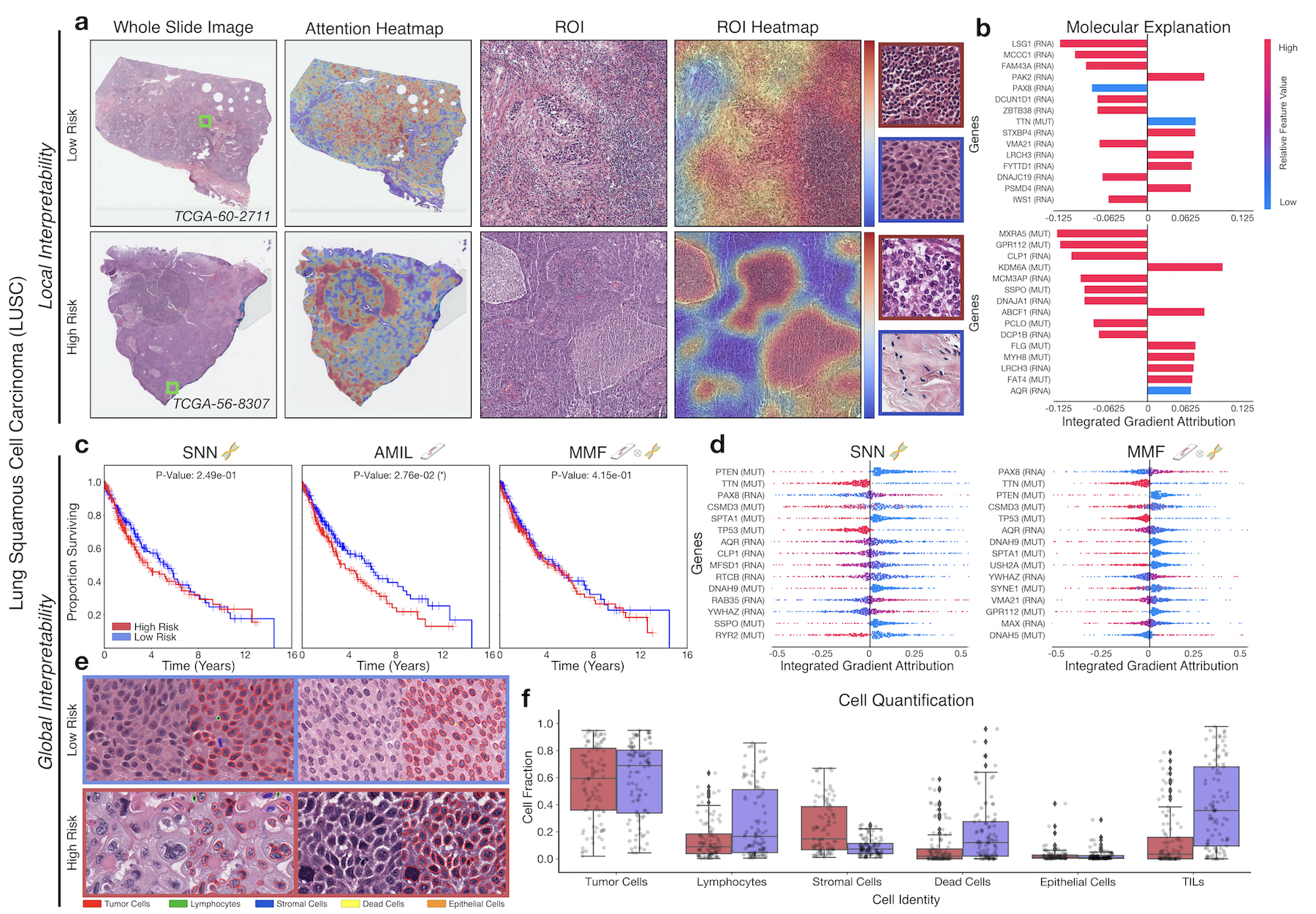}
\caption*{\textbf{Fig. S7: Quantitative performance, local model explanation, and global interpretability analyses of PORPOISE on LUSC.} \textbf{a.} WSIs, associated attention heatmaps, ROIs, ROI heatmaps, and selected high-attention patches from example low-risk (top) and high-risk (bottom) cases. In LUSC, high attention for low-risk cases tends to focus on regions with dense inflammatory infiltrate, predominantly comprised of lymphocytes (presumed TILs), while in high-risk cases, high attention focuses on regions of central necrosis within tumor nests. \textbf{b.} Local gene attributions for the corresponding low-risk (top) and high-risk (bottom) cases. \textbf{c.} Kaplan–Meier curves for omics-only (left, "SNN"), histology-only (center, "AMIL") and multimodal fusion (right, "MMF"), showing improved patient stratification over AMIL and late-stage patients in SNN. \textbf{d.} Global gene attributions across patient cohorts according to unimodal interpretability (left, "SNN"), and multimodal interpretability (right, "MMF"). \textbf{e.} High attention patches from low-risk (top) and high-risk (bottom) cases with corresponding cell labels. \textbf{f.} Quantification of cell types in high-attention patches for each disease overall, showing increased lymphocyte and TIL presence in low risk patients.}
\end{figure*}
\clearpage

\begin{figure*}
\vspace{-9mm}
\includegraphics[width=\textwidth]{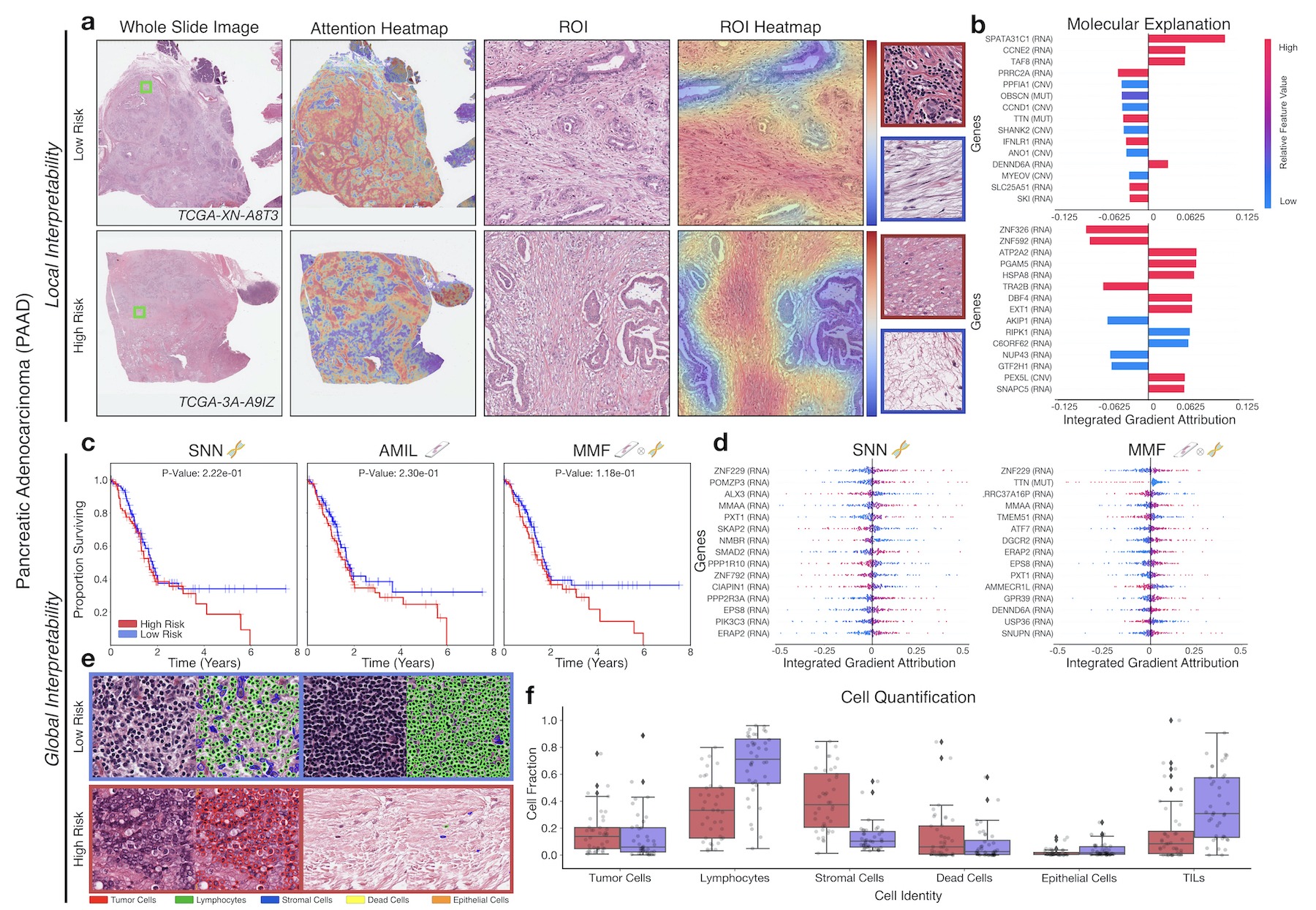}
\caption*{\textbf{Fig. S8: Quantitative performance, local model explanation, and global interpretability analyses of PORPOISE on PAAD.} \textbf{a.} WSIs, associated attention heatmaps, ROIs, ROI heatmaps, and selected high-attention patches from example low-risk (top) and high-risk (bottom) cases. In PAAD, high attention for low-risk cases tends to focus on stroma-contained dispersed glands and aggregates of lymphocytes, while in high-risk cases, high attention focuses on tumor-associated and myxoid stroma. \textbf{b.} Local gene attributions for the corresponding low-risk (top) and high-risk (bottom) cases. \textbf{c.} Kaplan–Meier curves for omics-only (left, "SNN"), histology-only (center, "AMIL") and multimodal fusion (right, "MMF"). \textbf{d.} Global gene attributions across patient cohorts according to unimodal interpretability (left, "SNN"), and multimodal interpretability (right, "MMF"). \textbf{e.} High attention patches from low-risk (top) and high-risk (bottom) cases with corresponding cell labels. \textbf{f.} Quantification of cell types in high-attention patches for each disease overall, showing increased lymphocyte and TIL presence in low risk patients.}
\end{figure*}
\clearpage

\begin{figure*}
\vspace{-9mm}
\includegraphics[width=\textwidth]{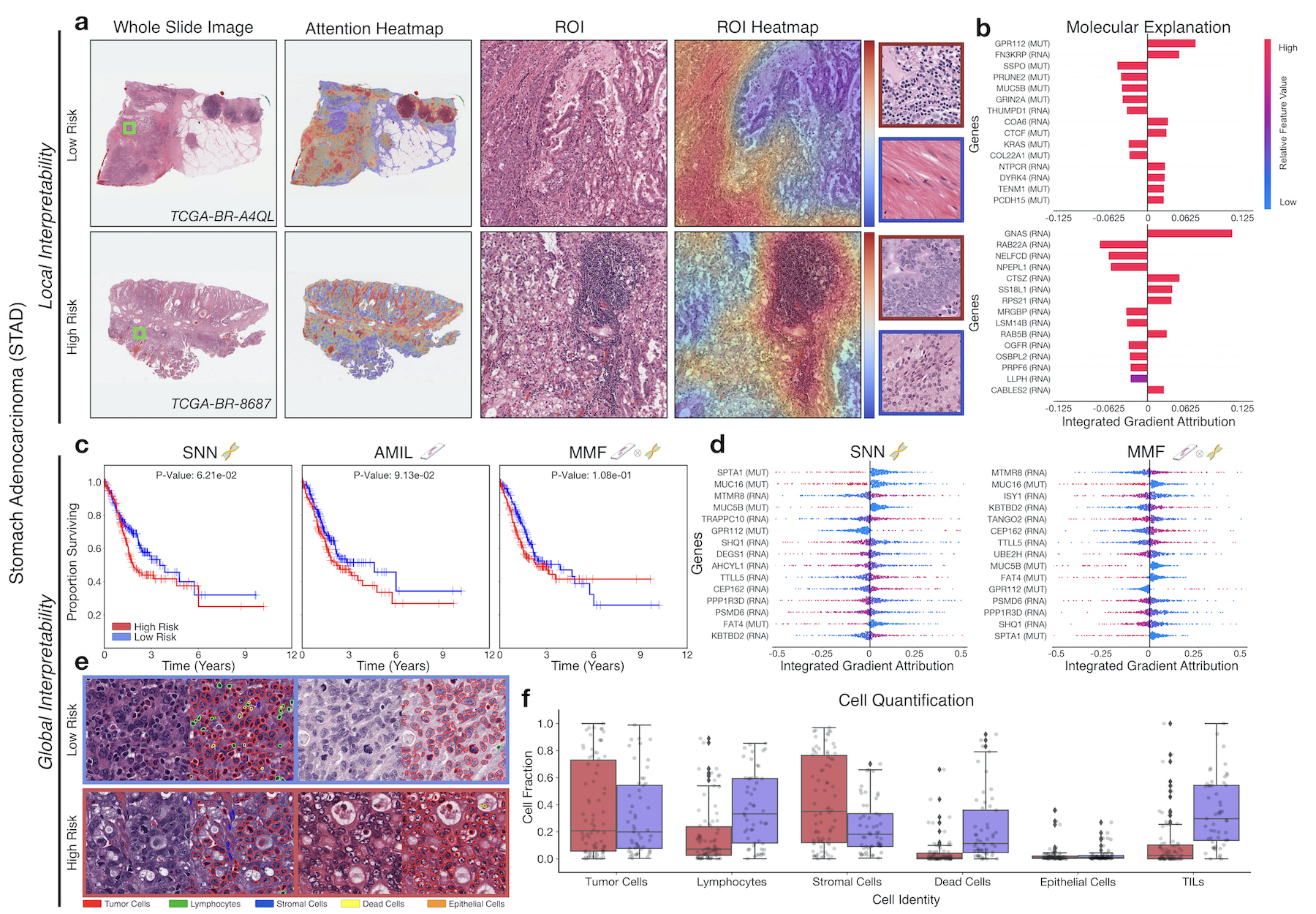}
\caption*{\textbf{Fig. S9: Quantitative performance, local model explanation, and global interpretability analyses of PORPOISE on STAD.} \textbf{a.} WSIs, associated attention heatmaps, ROIs, ROI heatmaps, and selected high-attention patches from example low-risk (top) and high-risk (bottom) cases. In STAD, high attention for low-risk cases tends to focus on dense regions of tumor, lymphocytes, and muscularis, while in high-risk cases, high attention focuses on dense regions of tumor and lymphocytes. \textbf{b.} Local gene attributions for the corresponding low-risk (top) and high-risk (bottom) cases. \textbf{c.} Kaplan–Meier curves for omics-only (left, "SNN"), histology-only (center, "AMIL") and multimodal fusion (right, "MMF"), showing better patient stratification in AMIL and SNN. \textbf{d.} Global gene attributions across patient cohorts according to unimodal interpretability (left, "SNN"), and multimodal interpretability (right, "MMF"). \textbf{e.} High attention patches from low-risk (top) and high-risk (bottom) cases with corresponding cell labels. \textbf{f.} Quantification of cell types in high-attention patches for each disease overall, showing XXX.}
\end{figure*}
\clearpage

\begin{figure*}
\vspace{-9mm}
\includegraphics[width=\textwidth]{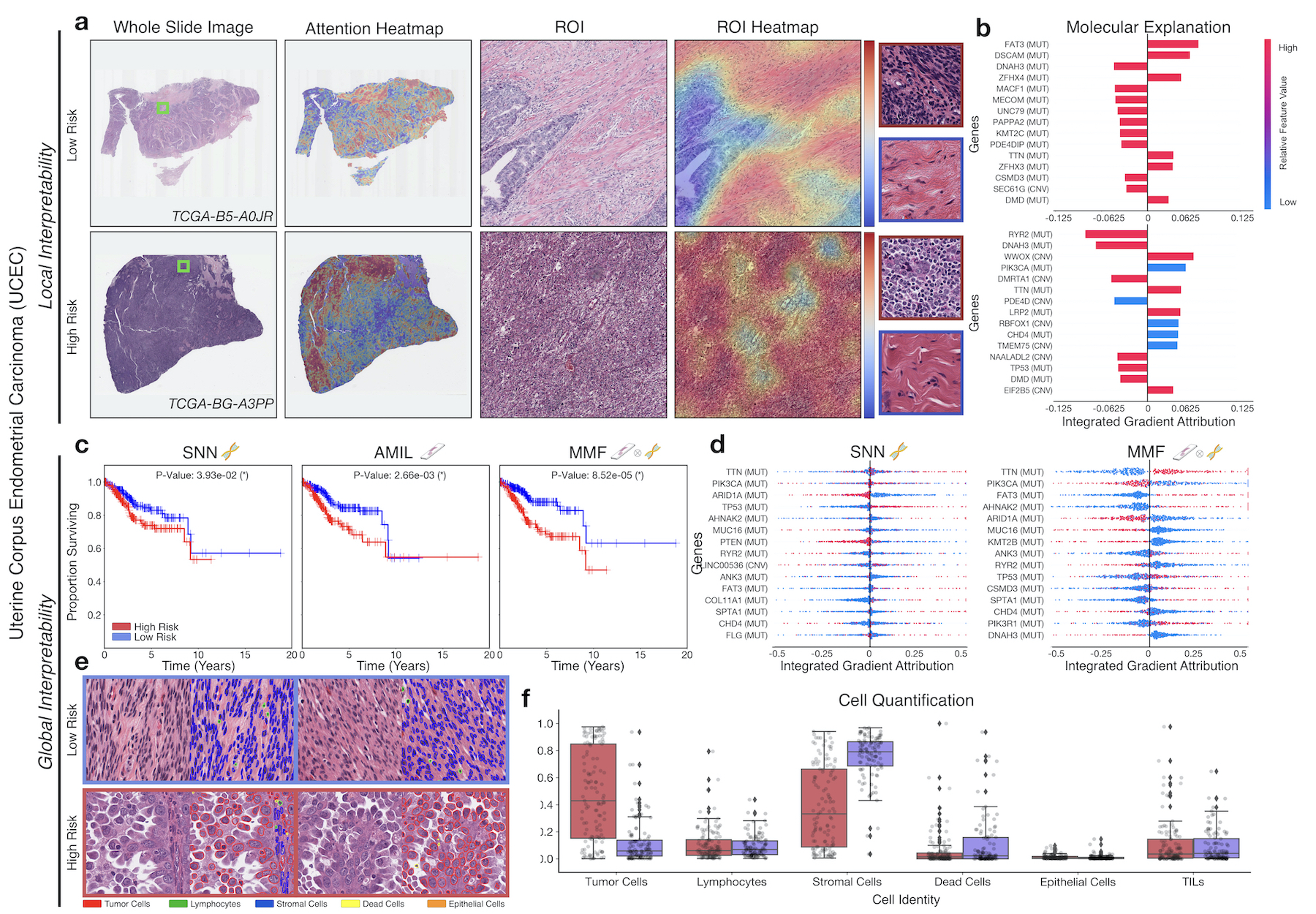}
\caption*{\textbf{Fig. S10: Quantitative performance, local model explanation, and global interpretability analyses of PORPOISE on UCEC.} \textbf{a.} WSIs, associated attention heatmaps, ROIs, ROI heatmaps, and selected high-attention patches from example low-risk (top) and high-risk (bottom) cases. In UCEC, high attention for low-risk cases tends to focus on background myometrium, while in high-risk cases, high attention focuses on tumor regions, especially those with increased nuclear pleomorphism and atypia. \textbf{b.} Local gene attributions for the corresponding low-risk (top) and high-risk (bottom) cases. \textbf{c.} Kaplan–Meier curves for omics-only (left, "SNN"), histology-only (center, "AMIL") and multimodal fusion (right, "MMF"), showing improved patient stratification over AMIL and late-stage patients in SNN. \textbf{d.} Global gene attributions across patient cohorts according to unimodal interpretability (left, "SNN"), and multimodal interpretability (right, "MMF"). \textbf{e.} High attention patches from low-risk (top) and high-risk (bottom) cases with corresponding cell labels. \textbf{f.} Quantification of cell types in high-attention patches for each disease overall, showing increased tumor cell presence in high risk patients and increased stromal cell presence in low risk patients.}
\end{figure*}
\clearpage

\begin{figure*}
\vspace{-9mm}
\includegraphics[width=\textwidth]{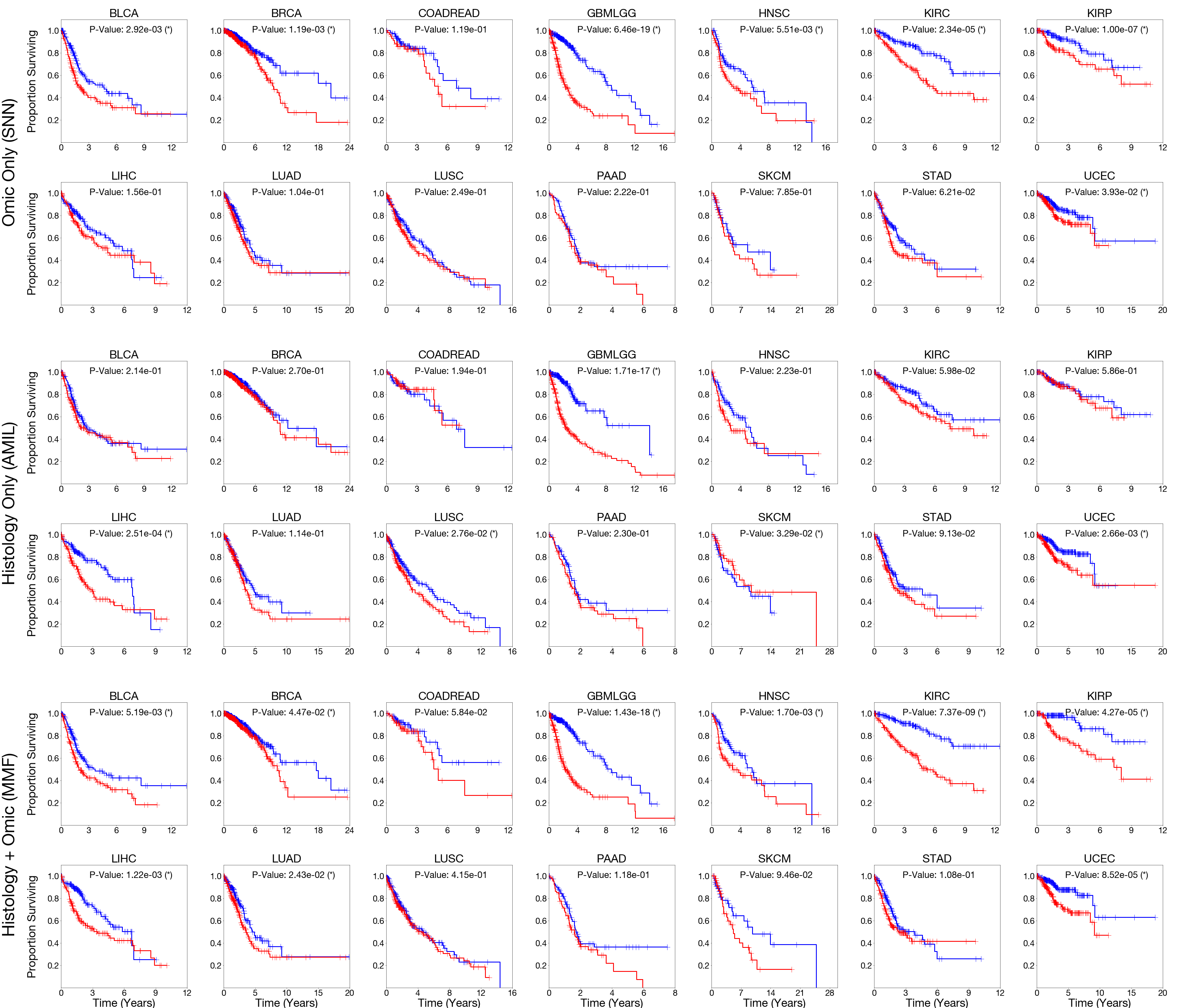}
\caption*{\textbf{Fig. S11: Kaplan-Meier comparative analysis of SNN, AMIL, and MMF.} Kaplan-Meier comparative analysis of SNN, AMIL, and MMF of patient stratification of low and high risk patients across all 14 cancer types. Low and high risks are defined by the median 50\% percentile of risk predictions. Logrank test was used to test for statistical significance in survival distributions between low and high risk patients (with * marked if P-Value $<$ 0.05).
}
\end{figure*}
\clearpage

\begin{figure*}
\vspace{-9mm}
\includegraphics[width=\textwidth]{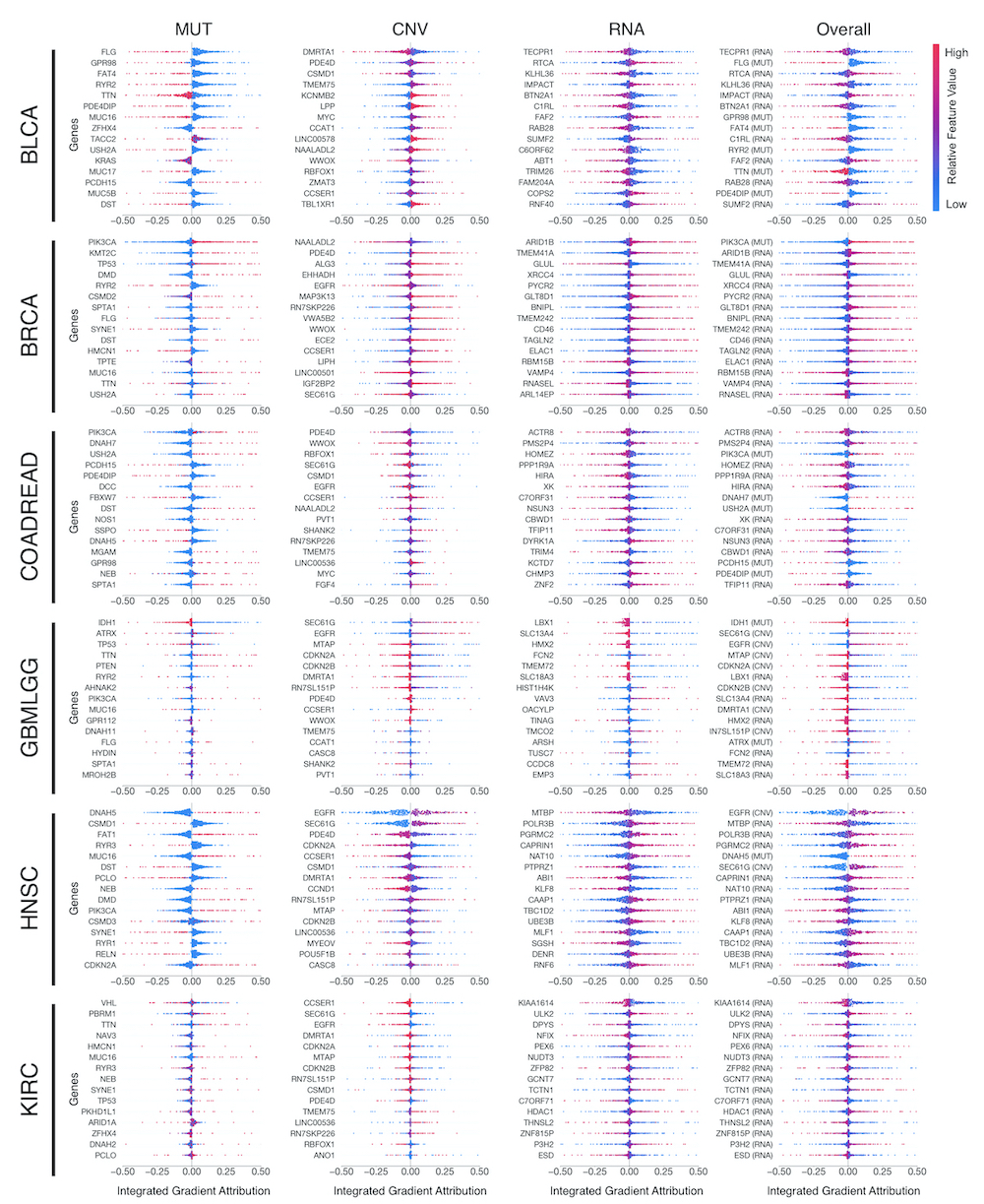}
\end{figure*}
\begin{figure*}
\vspace{-9mm}
\includegraphics[width=\textwidth]{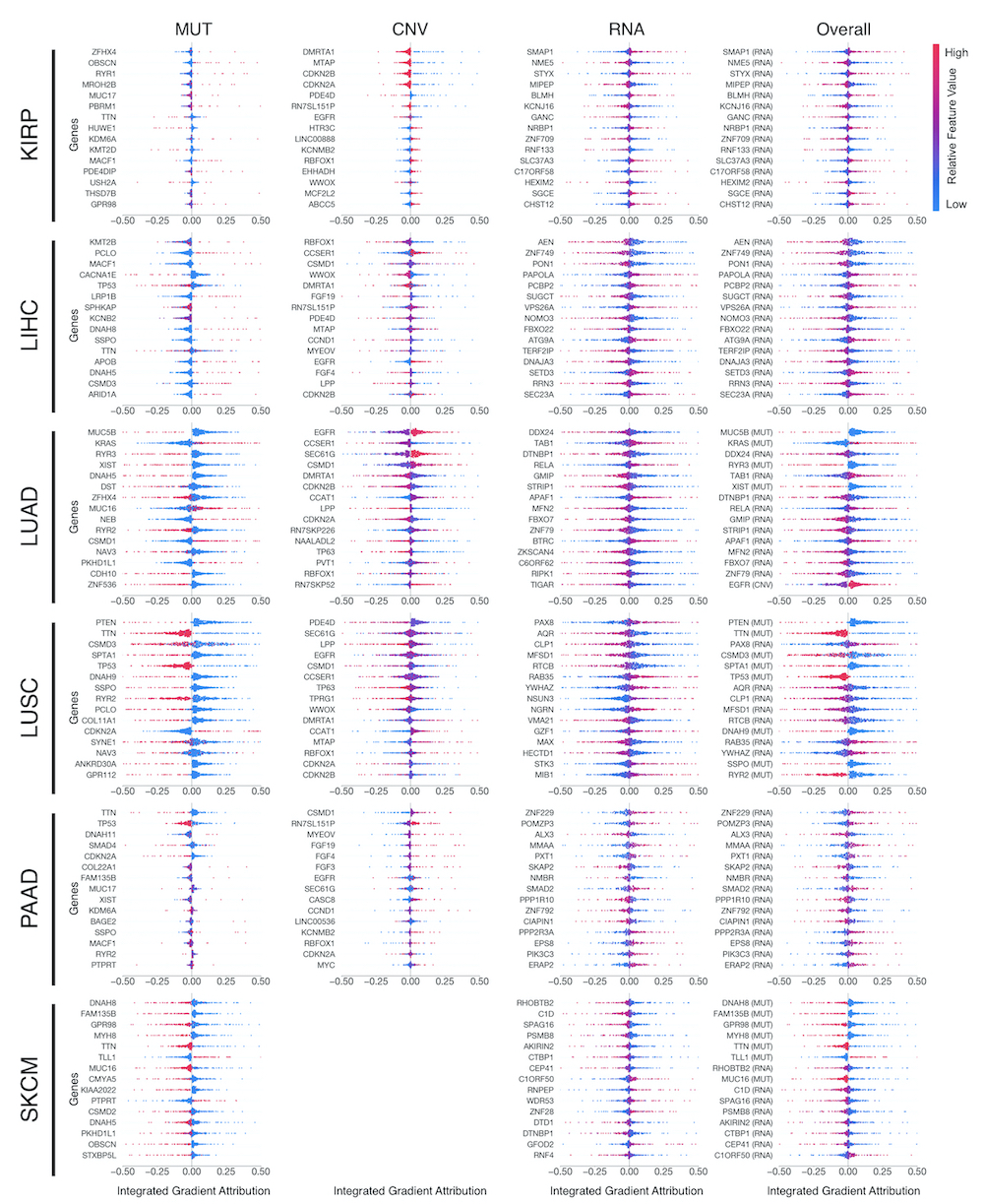}
\end{figure*}
\begin{figure*}
\vspace{-9mm}
\includegraphics[width=\textwidth]{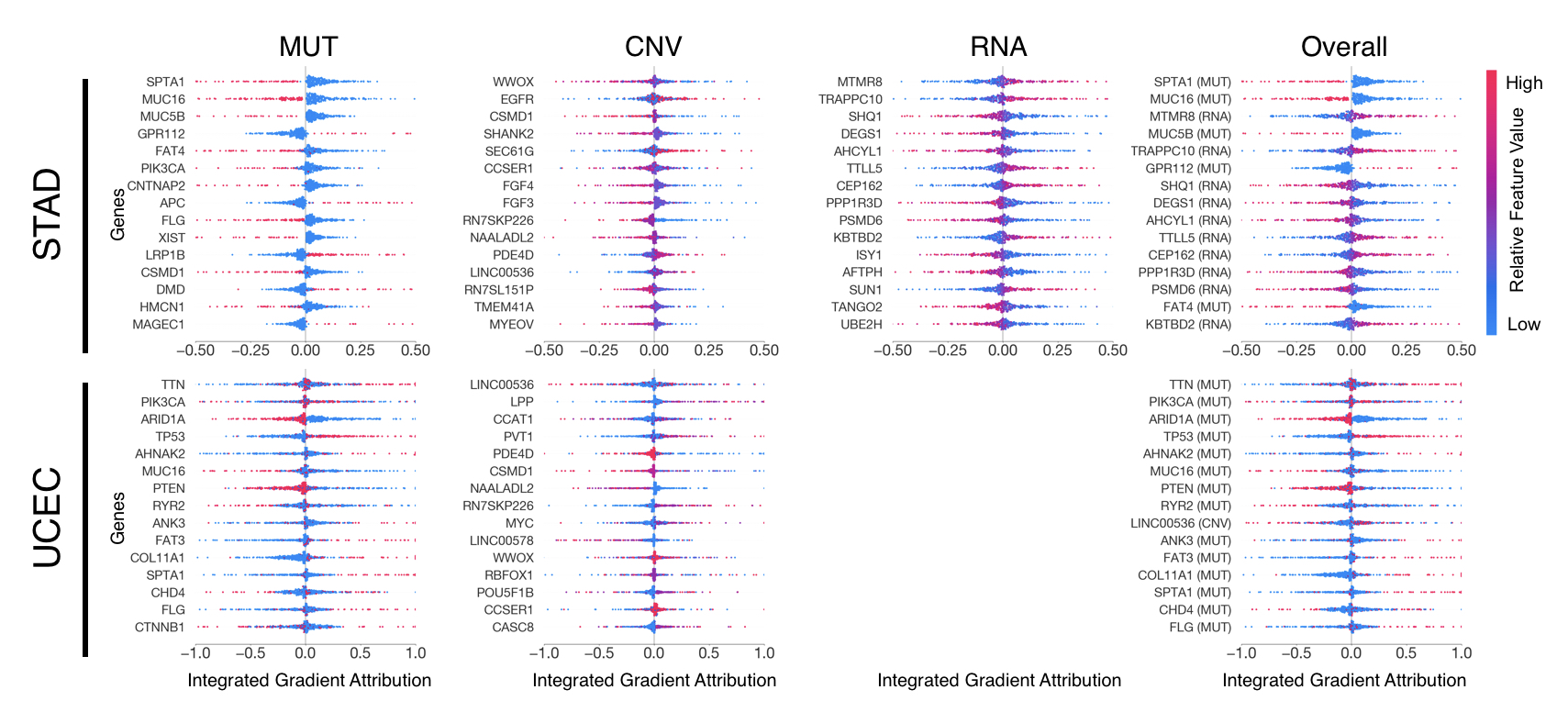}
\caption*{\textbf{Fig. S12: Global explanation of molecular features via SNN across cancer types.} Decision summary plots in each row illustrate top-15 molecular feature importances and their impact on cancer prognosis with respect to all patients in each cancer type. Molecular feature importances were computed using Integrated Gradients, and are separated by -omic data type (gene mutation status, gene copy number variation, RNA-Seq abundance). Attribution color corresponds to low (blue) vs. high (red) gene feature value, and attribution direction corresponds to how the gene feature value impacts low risk predictions (left) vs. high risk predictions (right). Data points in each summary plots correspond to local explanations made by Integrated Gradients when attributing features for a given sample. Top 15 features across each -omic data type were ranked by mean absolute attribution. SKCM CNV and UCEC RNA decision summary plots were missing due to missing CNV and RNA in SKCM and UCEC respectively.}
\end{figure*}
\clearpage

\begin{figure*}
\vspace{-9mm}
\includegraphics[width=\textwidth]{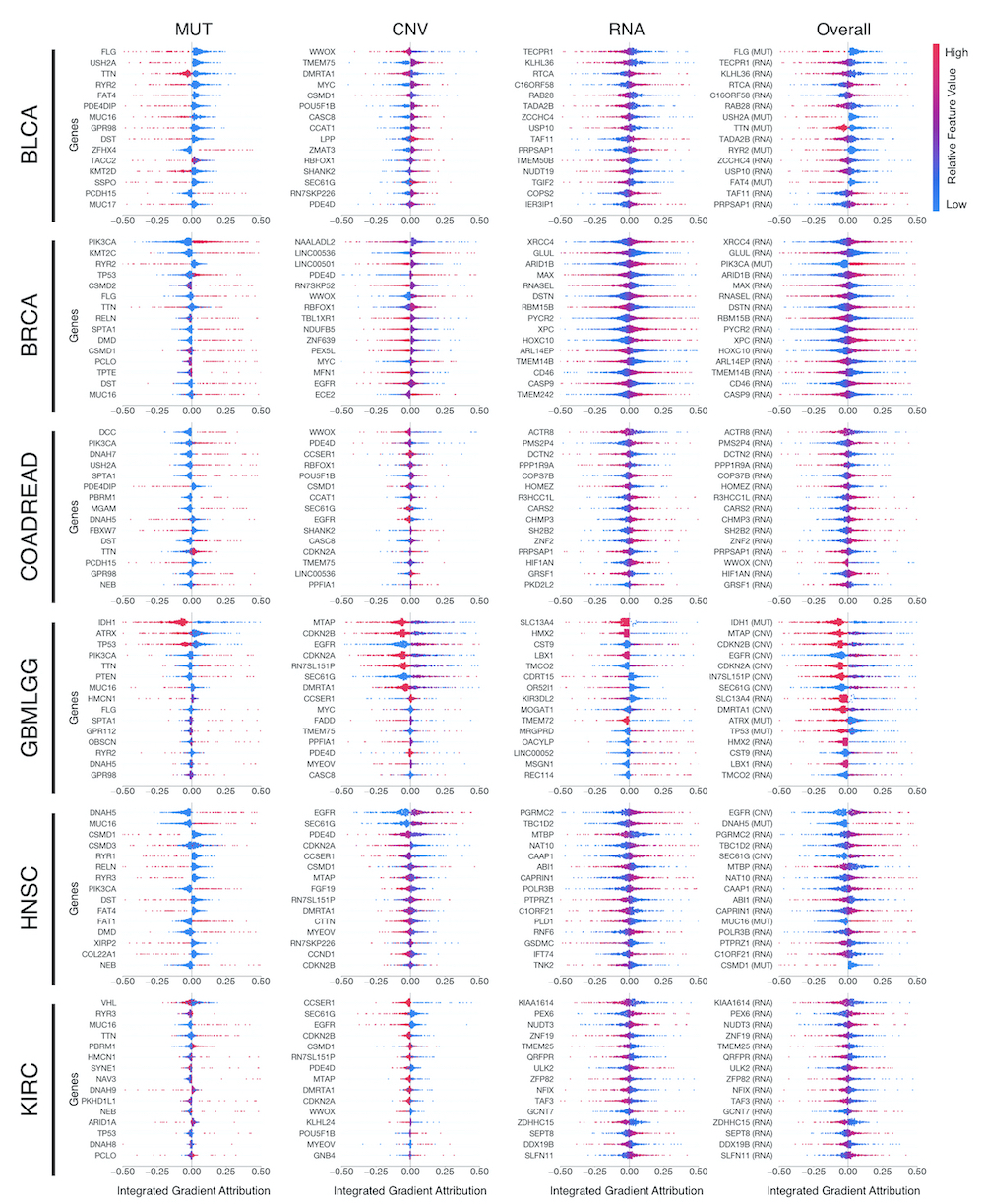}
\end{figure*}
\begin{figure*}
\vspace{-9mm}
\includegraphics[width=\textwidth]{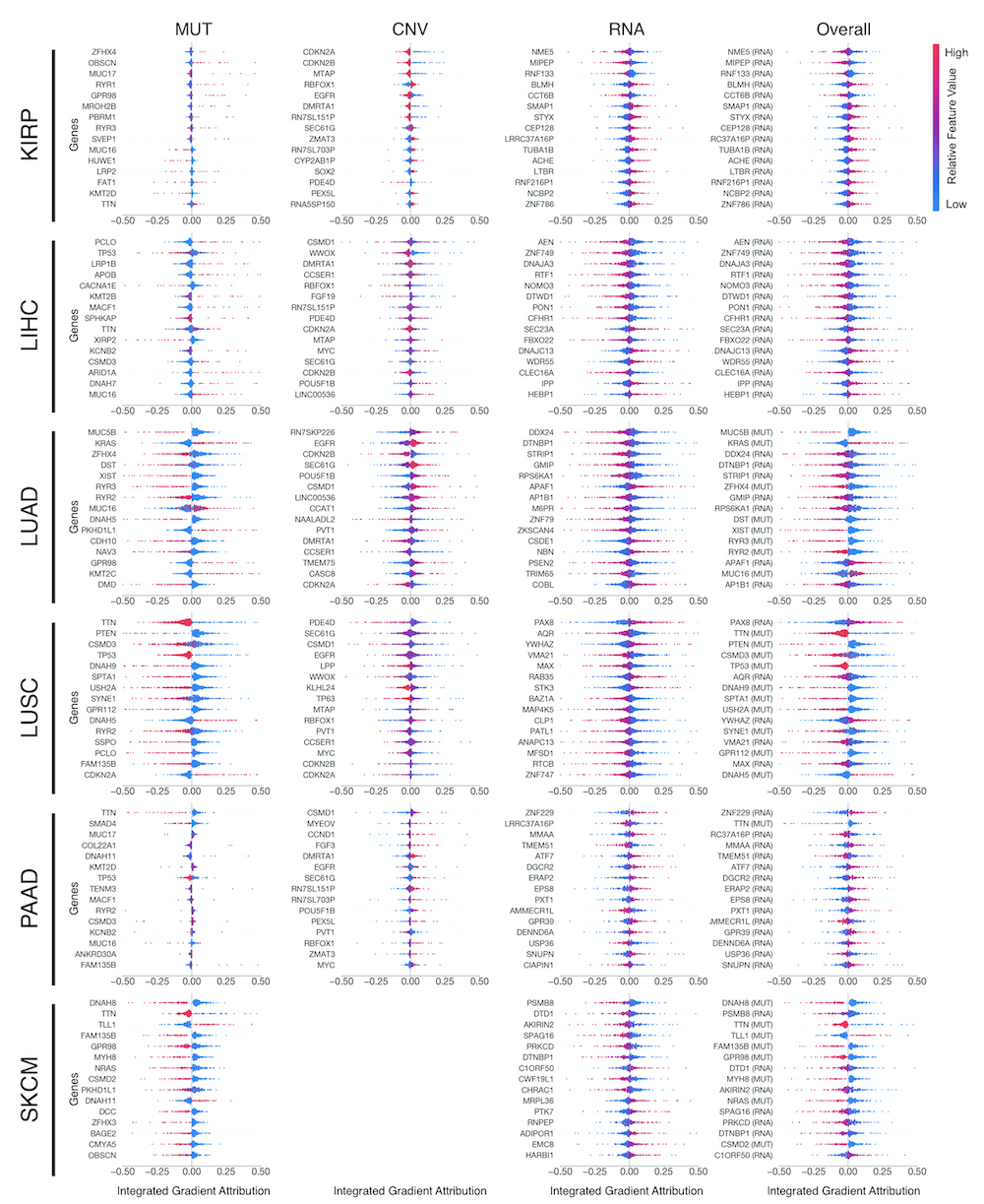}
\end{figure*}
\begin{figure*}
\vspace{-9mm}
\includegraphics[width=\textwidth]{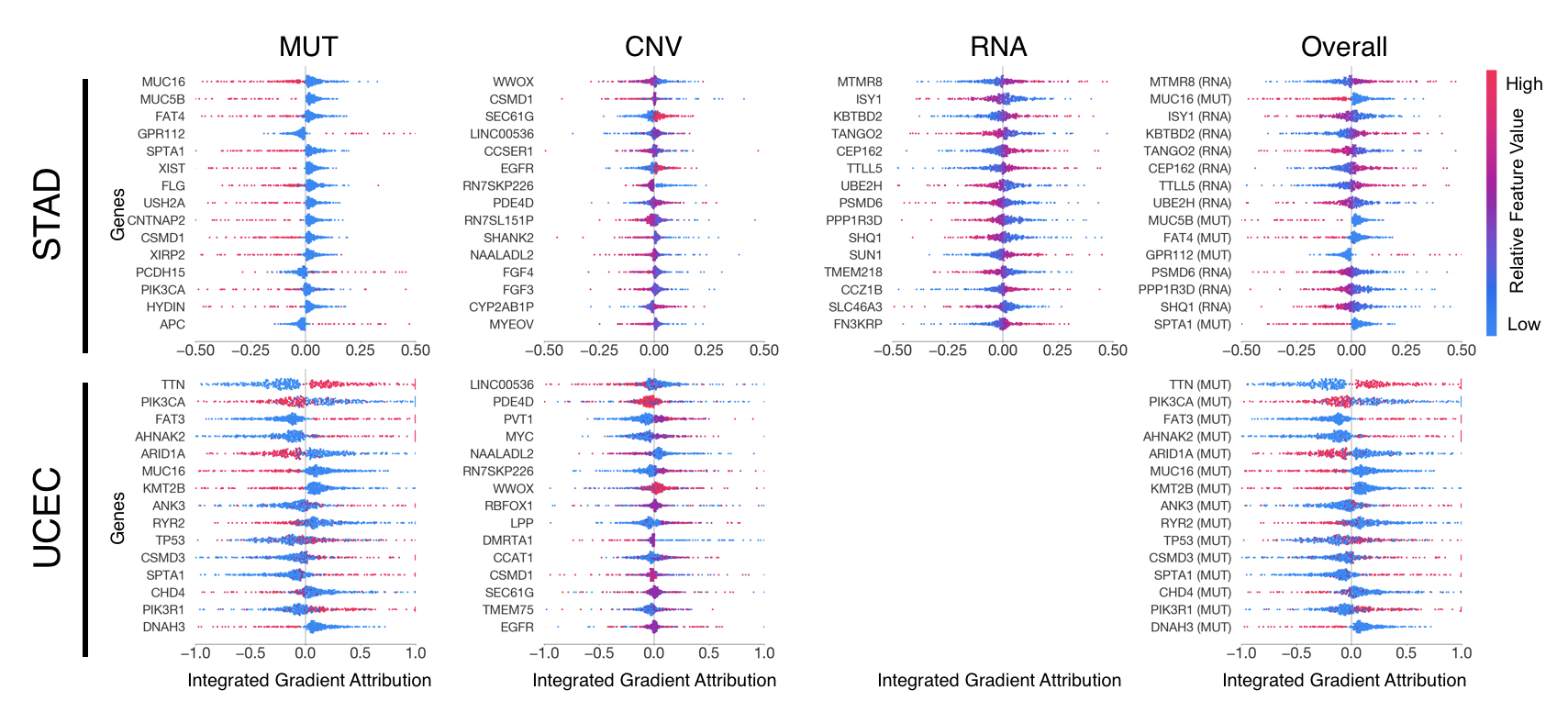}
\caption*{\textbf{Fig. S13: Global explanation of molecular features via MMF across cancer types.} Similar to Fig. S11, decision summary plots in each row illustrate top-15 molecular feature importances and their impact on cancer prognosis with respect to all patients in each cancer type. Similar to SNN, molecular feature importances were computed using Integrated Gradients, and are separated by -omic data type (gene mutation status, gene copy number variation, RNA-Seq abundance). SKCM CNV and UCEC RNA decision summary plots were missing due to missing CNV and RNA in SKCM and UCEC respectively.}
\end{figure*}
\clearpage

\begin{figure*}
\vspace{-9mm}
\includegraphics[width=\textwidth]{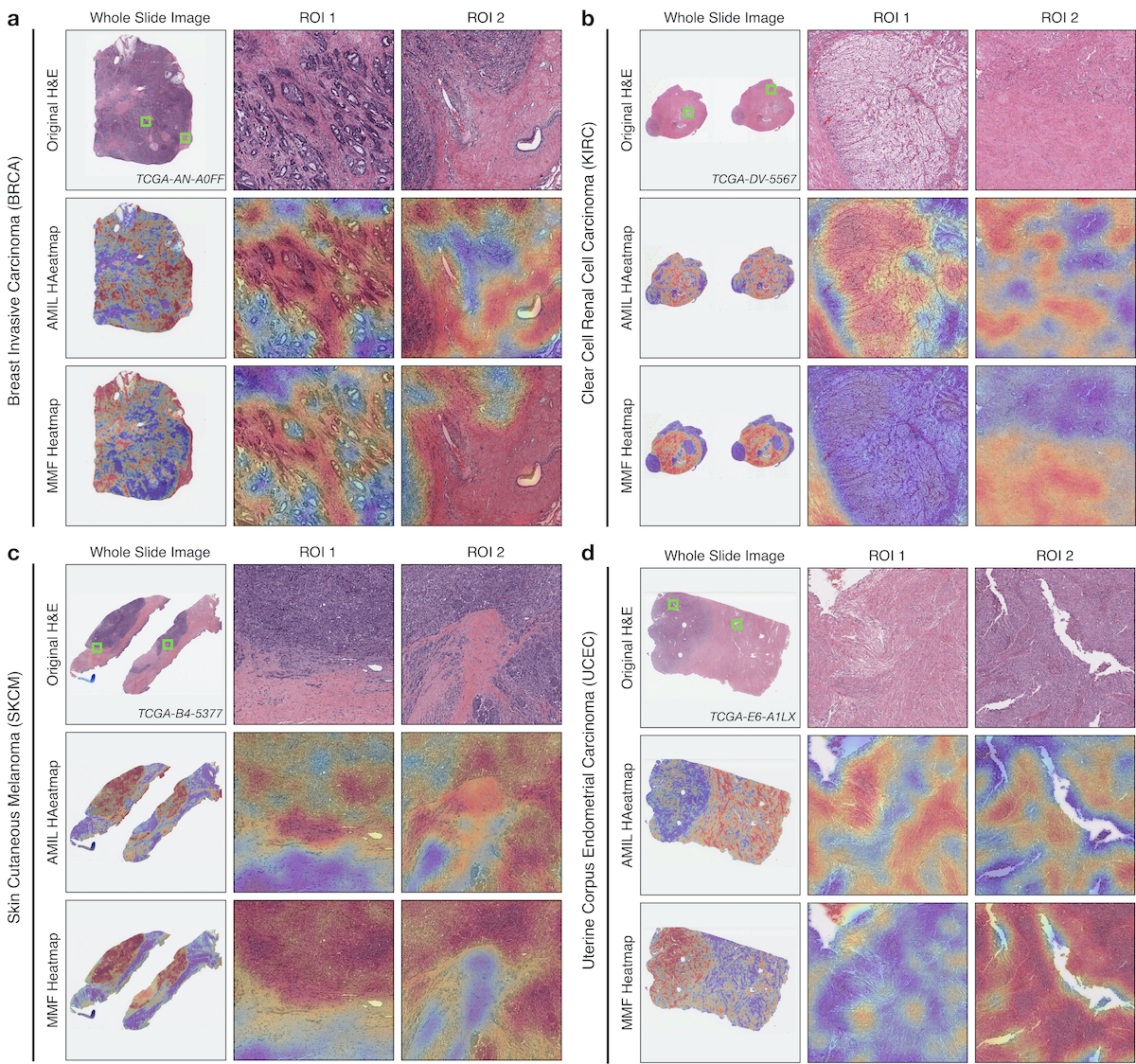}
\caption*{\textbf{Fig. S14: Exemplars of attention shift from unimodal to multimodal interpretability in WSIs.} Using PORPOISE, we investigated and examined how feature importance in high attention regions shifts when comparing AMIL (trained with histology-only) vs. MMF (histology conditioned with molecular profiles). In the assessment of each cancer type, "ROI 1" corresponds to a region where attention decreased from AMIL to MMF, and "ROI 2" corresponds to a region where attention increased from AMIL to MMF. \textbf{a.} In BRCA, attention shifted away from dense areas of tumor to both tumor and stromal regions in MMF. \textbf{b.} In KIRC, both stroma and tumor regions with classic "chicken-wire" vasculature are present in high attention regions in AMIL, whereas MMF attends to only stroma and completely segments out the tumor regions. \textbf{c.} In SKCM, both AMIL and MMF were able to localize tumor regions, with MMF being able to identify clear tumor-stroma boundaries. \textbf{d.} In UCEC, attention shifted towards dense tumor regions and away from stroma in MMF.
}
\end{figure*}
\clearpage

\clearpage

\begin{center}
\begin{table}
\footnotesize
\centering
\caption*{\textbf{Table S1. TCGA Pan-Cancer Demographic Characteristics}}
\centering
\begin{tabular}{l|rrrrrr}
\toprule
Cancer Type &  No. Cases &          Age & Gender M/F & White/Black/Other/NA & Grade 1/2/3/4/NA & Censorship \\
\midrule
BLCA        &        437 &  68.0 $\pm$ 10.6 &    327/110 &              358/23/14/42 &     21/0/388/0/3 & 0.533 \\
BRCA        &       1022 &  58.2 $\pm$ 13.3 &    12/1010 &             715/166/83/58 &        NA & 0.866 \\
COADREAD    &        345 &  64.9 $\pm$ 13.0 &    186/159 &              246/59/27/13 &        NA & 0.793 \\
GBMLGG      &       1011 &  46.0 $\pm$ 14.6 &    614/397 &               897/49/3/62 &  0/249/265/596/0 & 0.450 \\
HNSC        &        437 &  61.4 $\pm$ 11.9 &    321/116 &              367/47/13/10 &  63/311/125/7/22 & 0.538 \\
KIRC        &        350 &  60.0 $\pm$ 12.0 &    222/128 &                285/52/6/7 &  14/230/207/78/8 & 0.669 \\
KIRP        &        284 &  61.8 $\pm$ 12.4 &     211/73 &              203/59/12/10 &        NA & 0.847 \\
LIHC        &        347 &  59.5 $\pm$ 13.1 &    229/118 &              176/14/8/149 &  55/180/124/13/5 & 0.646 \\
LUAD        &        515 &  65.7 $\pm$ 10.1 &    227/288 &               402/53/52/8 &        NA & 0.608 \\
LUSC        &        484 &   67.5 $\pm$ 8.5 &    366/118 &              344/31/100/9 &        NA & 0.566 \\
PAAD        &        180 &  64.2 $\pm$ 10.8 &      93/87 &                159/6/4/11 &     32/97/51/2/3 & 0.457 \\
SKCM        &        268 &  58.1 $\pm$ 16.6 &     174/94 &                 252/1/6/9 &        NA & 0.528 \\
STAD        &        374 &  65.1 $\pm$ 10.4 &    246/128 &              236/10/47/81 &   12/159/263/0/9 & 0.609 \\
UCEC        &        538 &  64.1 $\pm$ 10.9 &      0/538 &             349/127/31/31 &  99/122/316/11/0 & 0.848 \\
\bottomrule
\end{tabular}
\end{table}
\end{center}

\clearpage
\begin{table}
\caption*{\textbf{Table S2. TCGA Pan-Cancer Feature Alignment Summary}}
\centering
\begin{tabular}{l|rrrr|rrr}
\toprule
{} &   WSI &   CNV &   MUT &   RNA &  WSI+CNV+MUT &  WSI+MUT+RNA &   ALL \\
Cancer Type  &       &       &       &       &              &              &       \\
\midrule
BLCA     &   454 &   443 &   452 &   450 &          441 &          448 &   \textbf{437} \\
BRCA     &  1117 &  1098 &  1040 &  1112 &         1025 &         1037 &   \textbf{1022} \\
COADREAD &   595 &   578 &   521 &   360 &          507 &          347 &   \textbf{345} \\
GBMLGG   &  1692 &  1658 &  1402 &  1049 &         1382 &         1026 &   \textbf{1011} \\
HNSC     &   470 &   464 &   451 &   462 &          445 &          443 &   \textbf{437} \\
KIRC     &   517 &   509 &   357 &   514 &          352 &          355 &   \textbf{350} \\
KIRP     &   294 &   291 &   286 &   293 &          284 &          285 &   \textbf{284} \\
LIHC     &   373 &   366 &   359 &   367 &          352 &          354 &   \textbf{347} \\
LUAD     &   528 &   522 &   523 &   522 &          519 &          519 &   \textbf{515} \\
LUSC     &   505 &   502 &   489 &   503 &          486 &          487 &   \textbf{484} \\
PAAD     &   208 &   201 &   187 &   195 &          187 &          180 &   \textbf{180} \\
 SKCM     &   464 &   162 &   269 &   271 &          160 &          \textbf{268} &   160 \\
STAD     &   404 &   402 &   400 &   379 &          398 &          376 &   \textbf{374} \\
UCEC     &   565 &   555 &   545 &   178 &          \textbf{538} &          172 &   171 \\
\bottomrule
\end{tabular}
\end{table}

\clearpage

\begin{center}
\begin{table}
\footnotesize
\caption*{\textbf{Table S3. Model Performance on Survival Prediction across 14 Cancer Types}}
\centering
\begin{tabular}{l|ll|ll|ll}
\toprule
{} & \multicolumn{2}{c|}{SNN} & \multicolumn{2}{c|}{AMIL} & \multicolumn{2}{c}{MMF}  \\
Cancer Type &                  c-Index (95\% CI) &   P-Value &                   c-Index (95\% CI) &   P-Value &              c-Index (95\% CI) &   P-Value \\
\midrule
BLCA     &  \textbf{0.632} (0.588-0.677) &  2.92e-03 &  0.542 (0.486-0.580) &  2.14e-01 &  0.627 (0.574-0.669) &  5.19e-03 \\
BRCA     &  \textbf{0.573} (0.508-0.637) &  1.19e-03 &  0.560 (0.489-0.615) &  2.70e-01 &  0.558 (0.498-0.620) &  4.47e-02 \\
COADREAD &  0.563 (0.498-0.631) &  5.41e-01 &  0.546 (0.466-0.617) &  2.44e-01 &  \textbf{0.580} (0.507-0.652) &  2.04e-01 \\
GBMLGG   &  0.751 (0.711-0.785) &  6.46e-19 &  \textbf{0.763} (0.718-0.795) &  1.71e-17 &  0.763 (0.718-0.794) &  1.43e-18 \\
HNSC     &  0.577 (0.533-0.624) &  8.33e-02 &  0.564 (0.507-0.592) &  3.15e-01 &  \textbf{0.580} (0.532-0.620) &  4.47e-02 \\
KIRC     &  0.665 (0.607-0.721) &  2.34e-05 &  0.567 (0.508-0.650) &  5.98e-02 &  \textbf{0.711} (0.648-0.757) &  7.37e-09 \\
KIRP     &  0.707 (0.599-0.793) &  1.00e-07 &  0.539 (0.408-0.625) &  5.86e-01 &  \textbf{0.811} (0.697-0.874) &  4.27e-05 \\
LIHC     &  0.570 (0.497-0.622) &  1.56e-01 &  0.618 (0.563-0.684) &  2.51e-04 &  \textbf{0.640} (0.569-0.693) &  1.22e-03 \\
LUAD     &  \textbf{0.591} (0.537-0.638) &  1.04e-01 &  0.548 (0.489-0.597) &  1.14e-01 &  0.586 (0.526-0.637) &  2.43e-02 \\
LUSC     &  0.522 (0.475-0.567) &  2.49e-01 &  \textbf{0.561} (0.500-0.597) &  2.76e-02 &  0.527 (0.472-0.557) &  4.15e-01 \\
PAAD     &  0.537 (0.476-0.607) &  2.22e-01 &  0.580 (0.485-0.613) &  2.30e-01 &  \textbf{0.591} (0.510-0.637) &  1.18e-01 \\
SKCM     &  0.519 (0.443-0.588) &  9.40e-01 &  0.607 (0.509-0.661) &  7.35e-03 &  \textbf{0.608} (0.532-0.674) &  4.09e-01 \\
STAD     &  0.545 (0.490-0.601) &  6.21e-02 &  0.556 (0.494-0.598) &  9.13e-02 &  \textbf{0.587} (0.537-0.643) &  1.08e-01 \\
UCEC     &  0.601 (0.515-0.660) &  3.93e-02 &  0.638 (0.563-0.701) &  2.66e-03 &  \textbf{0.644} (0.564-0.693) &  8.52e-05 \\
\midrule
Overall & 0.597 & - & 0.585 (0.601) & - & \textbf{0.630} & - \\
\bottomrule
\end{tabular}
\end{table}
\end{center}

\clearpage

\begin{table*}
\footnotesize
\caption*{\textbf{Table S4. Cox Baselines on Survival Prediction across 14 Cancer Types}}
\centering
\begin{tabular}{l|ll|ll|ll}
\toprule
{} & \multicolumn{2}{c|}{Grade} & \multicolumn{2}{c|}{Age+Gender} & \multicolumn{2}{c}{Age+Gender+Grade}  \\
Cancer Type &                  c-Index (95\% CI) &   P-Value &                   c-Index (95\% CI) &   P-Value &              c-Index (95\% CI) &   P-Value \\
\midrule
BLCA     &  0.518 (0.460-0.558) &  7.57e-01 &  0.613 (0.563-0.649) &  1.25e-02 &  0.620 (0.569-0.655) &  3.64e-03 \\
BRCA     &  - &  - &  0.645 (0.564-0.704) &  5.40e-04 &  0.645 (0.564-0.704) &  5.40e-04 \\
COADREAD &  - &  - &  0.601 (0.517-0.657) &  6.08e-04 &  0.601 (0.517-0.657) &  6.08e-04 \\
GBMLGG   &  0.783 (0.735-0.811) &  1.17e-20 &  0.768 (0.728-0.797) &  2.20e-19 &  0.841 (0.808-0.861) &  1.79e-28 \\
HNSC     &  0.473 (0.418-0.509) &  2.93e-01 &  0.555 (0.505-0.599) &  7.36e-03 &  0.553 (0.503-0.596) &  1.47e-02 \\
KIRC     &  0.646 (0.583-0.711) &  6.68e-04 &  0.621 (0.555-0.680) &  1.67e-04 &  0.678 (0.619-0.738) &  3.28e-07 \\
KIRP      &  - &  - &  0.565 (0.452-0.651) &  1.25e-01 &  0.565 (0.452-0.651) &  1.25e-01 \\
LIHC     &  0.556 (0.469-0.588) &  9.15e-01 &  0.510 (0.436-0.551) &  3.25e-01 &  0.513 (0.450-0.562) &  6.33e-01 \\
LUAD      &  - &  - &  0.557 (0.492-0.598) &  2.60e-01 &  0.557 (0.492-0.598) &  2.60e-01 \\
LUSC      &  - &  - &  0.566 (0.517-0.604) &  2.20e-02 &  0.566 (0.517-0.604) &  2.20e-02 \\
PAAD     &  0.585 (0.519-0.640) &  6.76e-02 &  0.551 (0.475-0.602) &  1.39e-01 &  0.571 (0.492-0.622) &  1.18e-02 \\
STAD      &  - &  - &  0.569 (0.477-0.622) &  5.00e-02 &  0.569 (0.477-0.622) &  5.00e-02 \\
SKCM     &  0.574 (0.509-0.613) &  9.86e-03 &  0.549 (0.493-0.598) &  5.74e-02 &  0.580 (0.533-0.635) &  1.08e-02 \\
UCEC     &  0.651 (0.586-0.706) &  4.80e-06 &  0.613 (0.563-0.649) &  1.25e-02 &  0.667 (0.592-0.715) &  3.04e-05 \\
\midrule
Overall & 0.598 & - & 0.592 & - & 0.608 & - \\
\bottomrule
\end{tabular}
\end{table*}

\clearpage

\begin{table}
\caption*{\textbf{Table S5. Average WSI attribution across 14 Cancer Types}}
\centering
\begin{tabular}{ll}
\toprule
{} & Avg. WSI Attribution \\
Cancer Type &                         \\
\midrule
BLCA        &           0.056 $\pm$ 0.025 \\
BRCA        &           0.111 $\pm$ 0.048 \\
COADREAD    &           0.092 $\pm$ 0.036 \\
GBMLGG      &           0.183 $\pm$ 0.092 \\
HNSC        &           0.039 $\pm$ 0.018 \\
KIRC        &           0.080 $\pm$ 0.032 \\
KIRP        &           0.165 $\pm$ 0.059 \\
LIHC        &           0.102 $\pm$ 0.043 \\
LUAD        &           0.056 $\pm$ 0.025 \\
LUSC        &           0.047 $\pm$ 0.018 \\
PAAD        &           0.083 $\pm$ 0.040 \\
STAD        &           0.070 $\pm$ 0.031 \\
SKCM        &           0.143 $\pm$ 0.052 \\
UCEC        &           0.175 $\pm$ 0.080 \\
\midrule
Overall & 0.100 \\
\bottomrule
\end{tabular}
\end{table}

\end{document}